\documentclass[12pt,a4paper]{article}

\usepackage[margin=1in]{geometry}
\usepackage{graphicx}
\usepackage{booktabs}
\usepackage{array}
\usepackage{cite}   
\usepackage{hyperref}
\usepackage{xcolor}
\usepackage{amsmath}
\usepackage{amssymb}
\usepackage{enumitem}
\usepackage{float}
\usepackage{caption}
\usepackage{subcaption}
\usepackage{tikz}
\usetikzlibrary{positioning, arrows.meta, fit, calc, backgrounds}

\definecolor{cData}{HTML}{DCE7F2}   \definecolor{cDataE}{HTML}{4A7BA7}
\definecolor{cExt}{HTML}{F4DDE2}    \definecolor{cExtE}{HTML}{A9556B}
\definecolor{cPrep}{HTML}{EEF1F4}   \definecolor{cRule}{HTML}{9AA3AB}
\definecolor{cModel}{HTML}{FAE6D5}  \definecolor{cModelE}{HTML}{B5763E}
\definecolor{cClass}{HTML}{E4EEE4}  \definecolor{cClassE}{HTML}{55804F}
\definecolor{cEval}{HTML}{E3ECE6}   \definecolor{cEvalE}{HTML}{3F6B55}
\definecolor{cCtrl}{HTML}{ECE7F0}   \definecolor{cCtrlE}{HTML}{6E5A82}
\usepackage{authblk}
\usepackage{microtype}  
\usepackage{titlesec}
\providecommand{\PaperValue}[1]{\csname paperonevalue@#1\endcsname}
\expandafter\def\csname paperonevalue@v006a6bb0055b1c0b\endcsname{11.1}
\expandafter\def\csname paperonevalue@v010139c57367b67f\endcsname{0.787}
\expandafter\def\csname paperonevalue@v0280292fe6c4e715\endcsname{0.035}
\expandafter\def\csname paperonevalue@v02ded317a0df5421\endcsname{0.507}
\expandafter\def\csname paperonevalue@v02e78d6b43b2c0cb\endcsname{9{,}712}
\expandafter\def\csname paperonevalue@v03ad6fa66a566705\endcsname{0.739}
\expandafter\def\csname paperonevalue@v0492ae6fcff07eb5\endcsname{3.0}
\expandafter\def\csname paperonevalue@v053fbf33f66c2937\endcsname{0.709}
\expandafter\def\csname paperonevalue@v05dcb455ceb71af7\endcsname{1.9}
\expandafter\def\csname paperonevalue@v061e2cacf78bd76c\endcsname{9.62}
\expandafter\def\csname paperonevalue@v068a2220460adc9a\endcsname{2{,}996}
\expandafter\def\csname paperonevalue@v07acc710b474d17e\endcsname{20.7}
\expandafter\def\csname paperonevalue@v089d14b5891f734c\endcsname{23}
\expandafter\def\csname paperonevalue@v0a71f280022bee49\endcsname{0.505}
\expandafter\def\csname paperonevalue@v0c7055b3ffd56f85\endcsname{44}
\expandafter\def\csname paperonevalue@v0e5c4529624f7407\endcsname{10}
\expandafter\def\csname paperonevalue@v0ed9f62fb1f14c3a\endcsname{13.8}
\expandafter\def\csname paperonevalue@v105e6aae0134da16\endcsname{0.565}
\expandafter\def\csname paperonevalue@v122f51ecfee0022f\endcsname{9.6}
\expandafter\def\csname paperonevalue@v1554fd9059e50c59\endcsname{58.1}
\expandafter\def\csname paperonevalue@v181e916396b84973\endcsname{19.3}
\expandafter\def\csname paperonevalue@v1879a2f5847674db\endcsname{0.490}
\expandafter\def\csname paperonevalue@v188e6eefc04fa197\endcsname{38{,}912}
\expandafter\def\csname paperonevalue@v19374cb3dd93f0e0\endcsname{0.0883}
\expandafter\def\csname paperonevalue@v1983b53706abb3bb\endcsname{79.3}
\expandafter\def\csname paperonevalue@v1bb724e9ce3cd967\endcsname{0.9998}
\expandafter\def\csname paperonevalue@v1bf24db5d16e37da\endcsname{7{,}284}
\expandafter\def\csname paperonevalue@v1d845c6999c3e8eb\endcsname{88}
\expandafter\def\csname paperonevalue@v1d999751296ac196\endcsname{4}
\expandafter\def\csname paperonevalue@v1ee1de8a99473f93\endcsname{7.8}
\expandafter\def\csname paperonevalue@v20a828c3dacaa344\endcsname{0.665}
\expandafter\def\csname paperonevalue@v21a4d0b9b1ca4e3b\endcsname{200}
\expandafter\def\csname paperonevalue@v226893c4a2e2fc65\endcsname{4.4}
\expandafter\def\csname paperonevalue@v2335a20175692296\endcsname{2}
\expandafter\def\csname paperonevalue@v23d9be02bce3f7b1\endcsname{0.670}
\expandafter\def\csname paperonevalue@v26cd9b1c2602fefe\endcsname{67.5}
\expandafter\def\csname paperonevalue@v277037d233fc1442\endcsname{1.000}
\expandafter\def\csname paperonevalue@v2a566e725669ebee\endcsname{5.4}
\expandafter\def\csname paperonevalue@v2c6439cdc1da1ac5\endcsname{173}
\expandafter\def\csname paperonevalue@v2c7ee9ace3685a7e\endcsname{87.8}
\expandafter\def\csname paperonevalue@v2c91484c325db6c8\endcsname{84.5}
\expandafter\def\csname paperonevalue@v2d9f829bf5835570\endcsname{0.0099}
\expandafter\def\csname paperonevalue@v2f1ff596853dc118\endcsname{1.000}
\expandafter\def\csname paperonevalue@v2f99e2ff8b4fe338\endcsname{66.1}
\expandafter\def\csname paperonevalue@v303bc19f9d64eedb\endcsname{0.090}
\expandafter\def\csname paperonevalue@v32bed4db5f25c7b7\endcsname{7.8}
\expandafter\def\csname paperonevalue@v32cdb6a650eac840\endcsname{8.2}
\expandafter\def\csname paperonevalue@v34432728441f3902\endcsname{0.5083}
\expandafter\def\csname paperonevalue@v353db98557019530\endcsname{70.0}
\expandafter\def\csname paperonevalue@v361a7d67b6215036\endcsname{3.2}
\expandafter\def\csname paperonevalue@v368139a4bff45772\endcsname{0.06}
\expandafter\def\csname paperonevalue@v37567c5e47c98348\endcsname{2}
\expandafter\def\csname paperonevalue@v392bd82ba0d864e3\endcsname{0.760}
\expandafter\def\csname paperonevalue@v3c8b7e4173713a09\endcsname{9.18}
\expandafter\def\csname paperonevalue@v3d569d488d4c156b\endcsname{0.0099}
\expandafter\def\csname paperonevalue@v3db52c932664da95\endcsname{17.5}
\expandafter\def\csname paperonevalue@v4033d5c509342cb0\endcsname{50}
\expandafter\def\csname paperonevalue@v4068a6ccfcce2bf7\endcsname{9.6}
\expandafter\def\csname paperonevalue@v420e51509322903f\endcsname{0.667}
\expandafter\def\csname paperonevalue@v429f8cb22b105cdb\endcsname{56.8}
\expandafter\def\csname paperonevalue@v45a4ef01b74d8208\endcsname{83.5}
\expandafter\def\csname paperonevalue@v46852ea18dccb863\endcsname{256}
\expandafter\def\csname paperonevalue@v47ad2d6406ec64ce\endcsname{2.8}
\expandafter\def\csname paperonevalue@v4857c3a67042bd6b\endcsname{0.1}
\expandafter\def\csname paperonevalue@v4b66988660e36874\endcsname{50}
\expandafter\def\csname paperonevalue@v4bead4fac078cdf3\endcsname{3}
\expandafter\def\csname paperonevalue@v4cd043e2f21b8269\endcsname{172}
\expandafter\def\csname paperonevalue@v4d41ce14edd9afcf\endcsname{0.671}
\expandafter\def\csname paperonevalue@v4eac869a6372d098\endcsname{1.000}
\expandafter\def\csname paperonevalue@v4f37e47574d60b71\endcsname{1.000}
\expandafter\def\csname paperonevalue@v5071b3c9d6a5451f\endcsname{1{,}187}
\expandafter\def\csname paperonevalue@v523603f3ff2e9f55\endcsname{0.7753}
\expandafter\def\csname paperonevalue@v546ff21179d07226\endcsname{0.789}
\expandafter\def\csname paperonevalue@v55eedafc6a50cc33\endcsname{0.043}
\expandafter\def\csname paperonevalue@v56b1335444acc45a\endcsname{3.5}
\expandafter\def\csname paperonevalue@v5743c89a1114e349\endcsname{0.687}
\expandafter\def\csname paperonevalue@v57817cece7ef7fe2\endcsname{4}
\expandafter\def\csname paperonevalue@v57ca568f41ac64da\endcsname{6.9}
\expandafter\def\csname paperonevalue@v57f1ab098fdca56f\endcsname{1.000}
\expandafter\def\csname paperonevalue@v58081b13c1e8f7fb\endcsname{82.3}
\expandafter\def\csname paperonevalue@v5819c9a7086222c3\endcsname{65}
\expandafter\def\csname paperonevalue@v59ed225df1912249\endcsname{0.794}
\expandafter\def\csname paperonevalue@v5a9d51e7b1ab59f4\endcsname{1.6}
\expandafter\def\csname paperonevalue@v5bba4711cb34a295\endcsname{0.010}
\expandafter\def\csname paperonevalue@v5bbcbf2399f05dcc\endcsname{9.4}
\expandafter\def\csname paperonevalue@v5bef026461f3d681\endcsname{67.1}
\expandafter\def\csname paperonevalue@v5c152b4002418033\endcsname{7.0}
\expandafter\def\csname paperonevalue@v5c716615cad2df62\endcsname{75.0}
\expandafter\def\csname paperonevalue@v5c7bfe3bf037c31a\endcsname{4.1}
\expandafter\def\csname paperonevalue@v5c83af940374dbbf\endcsname{1.1}
\expandafter\def\csname paperonevalue@v5c915601e2925637\endcsname{6.2}
\expandafter\def\csname paperonevalue@v5d9aa6cbb1c6f7dc\endcsname{0.677}
\expandafter\def\csname paperonevalue@v6049160f3660ec19\endcsname{0.012}
\expandafter\def\csname paperonevalue@v60d797a1ce397fa6\endcsname{0.563}
\expandafter\def\csname paperonevalue@v610bf82ff5236c83\endcsname{71.5}
\expandafter\def\csname paperonevalue@v62252131128de21a\endcsname{71.3}
\expandafter\def\csname paperonevalue@v62a7533244504e8a\endcsname{50}
\expandafter\def\csname paperonevalue@v6353b728a280369c\endcsname{21}
\expandafter\def\csname paperonevalue@v63effdd9a593ad1d\endcsname{75.3}
\expandafter\def\csname paperonevalue@v646bd7f009f3f4fd\endcsname{1.000}
\expandafter\def\csname paperonevalue@v650ae0ed5bcd6294\endcsname{0.7}
\expandafter\def\csname paperonevalue@v65e94ef249bf898d\endcsname{29}
\expandafter\def\csname paperonevalue@v66d94c8443fb746d\endcsname{1.000}
\expandafter\def\csname paperonevalue@v675121b0099df864\endcsname{0.772}
\expandafter\def\csname paperonevalue@v67f226f646217c2e\endcsname{6.3}
\expandafter\def\csname paperonevalue@v6810426a7cb2ffa4\endcsname{0.792}
\expandafter\def\csname paperonevalue@v6811171afff96c88\endcsname{72.0}
\expandafter\def\csname paperonevalue@v6843f6d2515ba41c\endcsname{20}
\expandafter\def\csname paperonevalue@v68560cf0795c08cb\endcsname{200}
\expandafter\def\csname paperonevalue@v68706bc58b0f31da\endcsname{100}
\expandafter\def\csname paperonevalue@v69e9d2d7bfd9fa79\endcsname{4.8}
\expandafter\def\csname paperonevalue@v6a420d0cc62641f0\endcsname{0.875}
\expandafter\def\csname paperonevalue@v6a6b32e1f7d0f2b5\endcsname{0.836}
\expandafter\def\csname paperonevalue@v6b8f67b2ee24abd4\endcsname{9.7}
\expandafter\def\csname paperonevalue@v6c06cce3e0a50704\endcsname{1.000}
\expandafter\def\csname paperonevalue@v6c4d910ddadcf793\endcsname{82.8}
\expandafter\def\csname paperonevalue@v6dd95784b595c020\endcsname{0.793}
\expandafter\def\csname paperonevalue@v6e48c0916b1d2d4f\endcsname{77.7}
\expandafter\def\csname paperonevalue@v6eba6c716a0e2e69\endcsname{68.1}
\expandafter\def\csname paperonevalue@v6ef6c05958f1383d\endcsname{0.498}
\expandafter\def\csname paperonevalue@v6f0f61a377b647ba\endcsname{0.18}
\expandafter\def\csname paperonevalue@v717677d42f14735f\endcsname{13.4}
\expandafter\def\csname paperonevalue@v729fba494eb3b3a0\endcsname{79.2}
\expandafter\def\csname paperonevalue@v733f49bd131c1794\endcsname{15.1}
\expandafter\def\csname paperonevalue@v7657968fc3d4d561\endcsname{0.805}
\expandafter\def\csname paperonevalue@v76bbf783959c8c12\endcsname{0.633}
\expandafter\def\csname paperonevalue@v779c63b5642ab262\endcsname{6.7}
\expandafter\def\csname paperonevalue@v780df097635c7608\endcsname{68.6}
\expandafter\def\csname paperonevalue@v79692b8e09fc9f85\endcsname{36}
\expandafter\def\csname paperonevalue@v7a1ed695d9e055b1\endcsname{53.7}
\expandafter\def\csname paperonevalue@v7b629439ea07f010\endcsname{0.835}
\expandafter\def\csname paperonevalue@v7d45acbdc13c4d10\endcsname{10.6}
\expandafter\def\csname paperonevalue@v7e372fd83049d9de\endcsname{0.691}
\expandafter\def\csname paperonevalue@v7f7936b17cbdaa1f\endcsname{74.7}
\expandafter\def\csname paperonevalue@v802caa2e93753a12\endcsname{11.9}
\expandafter\def\csname paperonevalue@v811a2c36bad919b3\endcsname{0.798}
\expandafter\def\csname paperonevalue@v815a3ee09aaf11c5\endcsname{100}
\expandafter\def\csname paperonevalue@v81ccfb305f7340cd\endcsname{9.4}
\expandafter\def\csname paperonevalue@v822c2b5c74b87d0e\endcsname{0.5653}
\expandafter\def\csname paperonevalue@v82c48b60d09b5c24\endcsname{0.488}
\expandafter\def\csname paperonevalue@v82fcef5335dde81a\endcsname{57}
\expandafter\def\csname paperonevalue@v85df81828ded8335\endcsname{0.068}
\expandafter\def\csname paperonevalue@v85f16b5dd393a698\endcsname{19}
\expandafter\def\csname paperonevalue@v863bbc5986a123e3\endcsname{4.6}
\expandafter\def\csname paperonevalue@v86bbfd61d5d0eb68\endcsname{14}
\expandafter\def\csname paperonevalue@v8724eed53aa08f27\endcsname{1.1}
\expandafter\def\csname paperonevalue@v8789cb06f501dc6c\endcsname{0.720}
\expandafter\def\csname paperonevalue@v87e7788a9c1fb692\endcsname{6.0}
\expandafter\def\csname paperonevalue@v881cf29a70ff4195\endcsname{0.5}
\expandafter\def\csname paperonevalue@v88a8596624cae846\endcsname{1.0}
\expandafter\def\csname paperonevalue@v88ff10ac5fe560fa\endcsname{64.6}
\expandafter\def\csname paperonevalue@v8a09eab4d99fd7df\endcsname{0.1}
\expandafter\def\csname paperonevalue@v8b57fcf7d74c901f\endcsname{59}
\expandafter\def\csname paperonevalue@v8b8e6dc8c36b6d62\endcsname{65}
\expandafter\def\csname paperonevalue@v8c344acbb3e6c42b\endcsname{0.6560}
\expandafter\def\csname paperonevalue@v8c7c2b99622623a6\endcsname{1.000}
\expandafter\def\csname paperonevalue@v8c87773002afea40\endcsname{18}
\expandafter\def\csname paperonevalue@v8c9ea0ac636080fb\endcsname{70.9}
\expandafter\def\csname paperonevalue@v8cc1fbf12d281798\endcsname{88}
\expandafter\def\csname paperonevalue@v8cedc6c358f1628c\endcsname{1.000}
\expandafter\def\csname paperonevalue@v8d266bb25d2ef367\endcsname{0.35}
\expandafter\def\csname paperonevalue@v8d68854096981149\endcsname{86.7}
\expandafter\def\csname paperonevalue@v8db5c489f89aa72c\endcsname{0.7074}
\expandafter\def\csname paperonevalue@v8e128931181e4de9\endcsname{23}
\expandafter\def\csname paperonevalue@v8f3e52e58a29c618\endcsname{70.0}
\expandafter\def\csname paperonevalue@v90b2db9ffaf5bc7a\endcsname{0.035}
\expandafter\def\csname paperonevalue@v91604c676610144e\endcsname{0.717}
\expandafter\def\csname paperonevalue@v91a0efa5161478aa\endcsname{0.7165}
\expandafter\def\csname paperonevalue@v91c163ee262cb289\endcsname{1.0000}
\expandafter\def\csname paperonevalue@v92029fbf6cc8451b\endcsname{44}
\expandafter\def\csname paperonevalue@v9297353d3eab8509\endcsname{0.669}
\expandafter\def\csname paperonevalue@v929e0bb0af4b5d3c\endcsname{0.114}
\expandafter\def\csname paperonevalue@v9309ff95441bded0\endcsname{13{,}170}
\expandafter\def\csname paperonevalue@v9354aab0e2577398\endcsname{42}
\expandafter\def\csname paperonevalue@v936f50c3c6559f35\endcsname{68.7}
\expandafter\def\csname paperonevalue@v943ac31c08c49693\endcsname{1.0}
\expandafter\def\csname paperonevalue@v94496782c36c58cc\endcsname{68.3}
\expandafter\def\csname paperonevalue@v9521ad1795eb7dc7\endcsname{0.502}
\expandafter\def\csname paperonevalue@v95de63f94c69e163\endcsname{0.36}
\expandafter\def\csname paperonevalue@v96559ce4c32d22e8\endcsname{6.9}
\expandafter\def\csname paperonevalue@v96a08164ca73fe56\endcsname{9.3}
\expandafter\def\csname paperonevalue@v98afc1e353c2581b\endcsname{0.488}
\expandafter\def\csname paperonevalue@v9a056397a7d45572\endcsname{0.8397}
\expandafter\def\csname paperonevalue@v9a7b56a1817bc304\endcsname{38{,}912}
\expandafter\def\csname paperonevalue@v9ad2a8acbb26ddba\endcsname{19}
\expandafter\def\csname paperonevalue@v9bfcf9c07ea7bc8e\endcsname{80.3}
\expandafter\def\csname paperonevalue@v9c0f67aae64c8b39\endcsname{76.7}
\expandafter\def\csname paperonevalue@v9c28514f5e8d4c99\endcsname{0.860}
\expandafter\def\csname paperonevalue@v9c86eb233cbc6ecb\endcsname{42}
\expandafter\def\csname paperonevalue@v9c8f20dc41e5355f\endcsname{23.4}
\expandafter\def\csname paperonevalue@v9dd56a66a888b39f\endcsname{7.3}
\expandafter\def\csname paperonevalue@v9e88bdaf265f0667\endcsname{200}
\expandafter\def\csname paperonevalue@v9f3e4a737fbd4f20\endcsname{0.781}
\expandafter\def\csname paperonevalue@va0d1ec09c13cec9d\endcsname{38{,}912}
\expandafter\def\csname paperonevalue@va0de40d901de51dc\endcsname{2{,}428}
\expandafter\def\csname paperonevalue@va1268b7d9f38d137\endcsname{1.5}
\expandafter\def\csname paperonevalue@va1c49c84f59c32c1\endcsname{7.0}
\expandafter\def\csname paperonevalue@va31d2d8d7fdc2ad8\endcsname{1.000}
\expandafter\def\csname paperonevalue@va341d39cd19b0dd6\endcsname{11{,}267}
\expandafter\def\csname paperonevalue@va4f8dff75a80ecd0\endcsname{0.736}
\expandafter\def\csname paperonevalue@va5132419b63691a3\endcsname{0.073}
\expandafter\def\csname paperonevalue@va5ff1f382f6b4e18\endcsname{0.793}
\expandafter\def\csname paperonevalue@va77236db5a776f0c\endcsname{0.007}
\expandafter\def\csname paperonevalue@va82a050cedcdcc24\endcsname{0.503}
\expandafter\def\csname paperonevalue@va9332b36af636f2b\endcsname{2{,}996}
\expandafter\def\csname paperonevalue@va964875514f7949e\endcsname{0.793}
\expandafter\def\csname paperonevalue@vac8d68626b06048b\endcsname{0.0198}
\expandafter\def\csname paperonevalue@vaca7da095d4a394c\endcsname{0.9991}
\expandafter\def\csname paperonevalue@vaccfdb0156dd60be\endcsname{0.0990}
\expandafter\def\csname paperonevalue@vada72137c6342650\endcsname{1.6}
\expandafter\def\csname paperonevalue@vade98305d10334de\endcsname{3000}
\expandafter\def\csname paperonevalue@vb19d0558443f5600\endcsname{83.6}
\expandafter\def\csname paperonevalue@vb2f9e30dc909bef7\endcsname{4.3}
\expandafter\def\csname paperonevalue@vb370fc4c5a7faa29\endcsname{0.793}
\expandafter\def\csname paperonevalue@vb5ae71193d8cb527\endcsname{0.038}
\expandafter\def\csname paperonevalue@vb8c001cb226923a9\endcsname{0.734}
\expandafter\def\csname paperonevalue@vb923d3d63229a821\endcsname{0.669}
\expandafter\def\csname paperonevalue@vb9eed640f212fafe\endcsname{40}
\expandafter\def\csname paperonevalue@vbb379e938607be31\endcsname{253}
\expandafter\def\csname paperonevalue@vbbe1e3b71628f266\endcsname{83.3}
\expandafter\def\csname paperonevalue@vbc319544810195ec\endcsname{70.2}
\expandafter\def\csname paperonevalue@vbcfa8f09aa392952\endcsname{0.61}
\expandafter\def\csname paperonevalue@vbdd01c1fb5051cbe\endcsname{0.767}
\expandafter\def\csname paperonevalue@vbf3bbb54944f7133\endcsname{79.3}
\expandafter\def\csname paperonevalue@vbfa278f28ef1e2b9\endcsname{1.7}
\expandafter\def\csname paperonevalue@vc0a8a954f825e359\endcsname{0.569}
\expandafter\def\csname paperonevalue@vc0d17f37b42efdb9\endcsname{0.752}
\expandafter\def\csname paperonevalue@vc1d63144fdb94693\endcsname{76.1}
\expandafter\def\csname paperonevalue@vc3f761bbb7600b12\endcsname{1{,}187}
\expandafter\def\csname paperonevalue@vc669d37f6b83ba1f\endcsname{15.1}
\expandafter\def\csname paperonevalue@vc6ca7300571d5b6e\endcsname{73.4}
\expandafter\def\csname paperonevalue@vc76775337296598a\endcsname{61.1}
\expandafter\def\csname paperonevalue@vc76e0557a3b9825e\endcsname{14.1}
\expandafter\def\csname paperonevalue@vc79915caa9a7bd27\endcsname{6.5}
\expandafter\def\csname paperonevalue@vc856505b802e9cda\endcsname{0.517}
\expandafter\def\csname paperonevalue@vc9b151715e9af58f\endcsname{18}
\expandafter\def\csname paperonevalue@vc9df4ac5d2795b7a\endcsname{6.3}
\expandafter\def\csname paperonevalue@vc9e75c1e5a8373f1\endcsname{0.7039}
\expandafter\def\csname paperonevalue@vcaec29a078df7ebc\endcsname{1.000}
\expandafter\def\csname paperonevalue@vcb0e73537a69568e\endcsname{142}
\expandafter\def\csname paperonevalue@vcb589cdcf41ef0be\endcsname{1.000}
\expandafter\def\csname paperonevalue@vcbbc4d939e299c96\endcsname{3{,}000}
\expandafter\def\csname paperonevalue@vcc64374fac1bbeb9\endcsname{64.3}
\expandafter\def\csname paperonevalue@vcd37033d8aead606\endcsname{0.106}
\expandafter\def\csname paperonevalue@vce6a66b1022fbfde\endcsname{0.506}
\expandafter\def\csname paperonevalue@vcf3c5516f2bc4883\endcsname{4.6}
\expandafter\def\csname paperonevalue@vcf5c85c16b40ab3d\endcsname{0.1}
\expandafter\def\csname paperonevalue@vd01f79f27741aa43\endcsname{8{,}130}
\expandafter\def\csname paperonevalue@vd04ca523a8326f16\endcsname{0.495}
\expandafter\def\csname paperonevalue@vd0d193f424fcfe36\endcsname{1.000}
\expandafter\def\csname paperonevalue@vd13157aa2eaff822\endcsname{242}
\expandafter\def\csname paperonevalue@vd29f27e84a2ae739\endcsname{0.527}
\expandafter\def\csname paperonevalue@vd2f673b26fed6ffb\endcsname{10.21}
\expandafter\def\csname paperonevalue@vd3860d9782b9539c\endcsname{200}
\expandafter\def\csname paperonevalue@vd48be45818bd4cba\endcsname{5.7}
\expandafter\def\csname paperonevalue@vd4d580014d7f8622\endcsname{14}
\expandafter\def\csname paperonevalue@vd4e1346b00fccff9\endcsname{4.7}
\expandafter\def\csname paperonevalue@vd5b9323303ec3f3f\endcsname{2.1}
\expandafter\def\csname paperonevalue@vd6052d609c05a63e\endcsname{42}
\expandafter\def\csname paperonevalue@vd6b8823cda21a851\endcsname{0.040}
\expandafter\def\csname paperonevalue@vd72c6bcd4096b637\endcsname{0.687}
\expandafter\def\csname paperonevalue@vd8a7795a4750560e\endcsname{1.000}
\expandafter\def\csname paperonevalue@vd9365d4b75df7fc6\endcsname{34}
\expandafter\def\csname paperonevalue@vd96fe4f01f5247e1\endcsname{87.5}
\expandafter\def\csname paperonevalue@vda05124b7f49ccbf\endcsname{0.0099}
\expandafter\def\csname paperonevalue@vda7643566ef0f63c\endcsname{66.9}
\expandafter\def\csname paperonevalue@vda8a8a64d03d9a6b\endcsname{9.6}
\expandafter\def\csname paperonevalue@vda95aa60647b6107\endcsname{0.0099}
\expandafter\def\csname paperonevalue@vdc0745a0fd565316\endcsname{237}
\expandafter\def\csname paperonevalue@vde9d729bb7152fbd\endcsname{1.000}
\expandafter\def\csname paperonevalue@ve1fd0c6fac5ac184\endcsname{0.5}
\expandafter\def\csname paperonevalue@ve2ca19956fedce4f\endcsname{68.6}
\expandafter\def\csname paperonevalue@ve3df06a4e857ec2a\endcsname{11.22}
\expandafter\def\csname paperonevalue@ve54ede0b638c1a02\endcsname{4{,}807}
\expandafter\def\csname paperonevalue@ve57fc765a91cb8dc\endcsname{62.3}
\expandafter\def\csname paperonevalue@ve65202e34fc89947\endcsname{0.506}
\expandafter\def\csname paperonevalue@ve78b44b95d50becb\endcsname{0.6250}
\expandafter\def\csname paperonevalue@ve7beb0d75f9145b2\endcsname{0.3}
\expandafter\def\csname paperonevalue@ve7e6246d845f1eb4\endcsname{67.7}
\expandafter\def\csname paperonevalue@ve80637b0015c0e39\endcsname{12.4}
\expandafter\def\csname paperonevalue@ve80dfe151b5c24b2\endcsname{1.9}
\expandafter\def\csname paperonevalue@ve88e1fe14712ab15\endcsname{10.2}
\expandafter\def\csname paperonevalue@ve8907089e5fdcc61\endcsname{80.6}
\expandafter\def\csname paperonevalue@ve8bc7acec10e28af\endcsname{52.7}
\expandafter\def\csname paperonevalue@ve9a03c4b0ed0ec4e\endcsname{15.5}
\expandafter\def\csname paperonevalue@ve9e7ee15373fc2d2\endcsname{1.0000}
\expandafter\def\csname paperonevalue@vea7ad4b657e4f2d6\endcsname{0.691}
\expandafter\def\csname paperonevalue@veb5cb2315c385ddf\endcsname{11.8}
\expandafter\def\csname paperonevalue@vebd7d8a600e609f8\endcsname{0.802}
\expandafter\def\csname paperonevalue@veca973b0ef491ef6\endcsname{0.578}
\expandafter\def\csname paperonevalue@vecc2f05045986fc1\endcsname{0.666}
\expandafter\def\csname paperonevalue@veda5a9d9896e7ed5\endcsname{0.537}
\expandafter\def\csname paperonevalue@veea8976c8b90d9bd\endcsname{0.696}
\expandafter\def\csname paperonevalue@vf0606d3c224b784c\endcsname{0.568}
\expandafter\def\csname paperonevalue@vf3452f892263c4e7\endcsname{18.3}
\expandafter\def\csname paperonevalue@vf369e6b562afdcd0\endcsname{2.9}
\expandafter\def\csname paperonevalue@vf405fe8ac2eac9d6\endcsname{3.2}
\expandafter\def\csname paperonevalue@vf40da55154906991\endcsname{20}
\expandafter\def\csname paperonevalue@vf4de53a42934ff68\endcsname{43}
\expandafter\def\csname paperonevalue@vf55a08d33c89b541\endcsname{0.8392}
\expandafter\def\csname paperonevalue@vf6bf55f887680628\endcsname{42}
\expandafter\def\csname paperonevalue@vf6faed7a5fbbafe6\endcsname{0.492}
\expandafter\def\csname paperonevalue@vf8e66e8a6ceffaac\endcsname{0.570}
\expandafter\def\csname paperonevalue@vf905612104b49878\endcsname{1.5}
\expandafter\def\csname paperonevalue@vfab46c67ff68f0b5\endcsname{200}
\expandafter\def\csname paperonevalue@vfc5dd21abc2243b1\endcsname{5.38}
\expandafter\def\csname paperonevalue@vfcb74078420c5ad6\endcsname{43}
\expandafter\def\csname paperonevalue@vfe75b9de5087c30e\endcsname{0.574}
\expandafter\def\csname paperonevalue@vfeda25c57cf14cea\endcsname{0.495}
\expandafter\def\csname paperonevalue@vff2b6b82a8c3e683\endcsname{2.9}

\titleformat{\section}[hang]
  {\normalfont\Large\bfseries\filright}{\thesection\hspace{0.75em}}{0pt}{}
\titleformat{\subsection}[hang]
  {\normalfont\large\bfseries\filright}{\thesubsection\hspace{0.75em}}{0pt}{}

\hypersetup{
  colorlinks=true,
  linkcolor=blue!60!black,
  urlcolor=blue!60!black,
  citecolor=blue!60!black,
  pdftitle={A Negative-Control Protocol for Clinical EEG Foundation-Model Benchmarks: Dataset Identity and External-Cohort Stress Testing},
  pdfauthor={Marzieh Zare}
}

\title{A Negative-Control Protocol for Clinical EEG Foundation-Model Benchmarks: Dataset Identity and External-Cohort Stress Testing}

\author{Marzieh Zare}
\affil{Universit\'{e} Laval, School of Psychology, Quebec City, QC, Canada}

\date{}

\begin{document}
\maketitle

\begin{abstract}

\noindent\textbf{Objective.} Apparent clinical EEG foundation-model gains may reflect cohort, montage, or probe design. We test whether these gains survive an external-cohort stress test and targeted negative controls.

\noindent\textbf{Approach.} We evaluated five pretrained encoders on five benchmark tasks and one Korean CAUEEG task, using model-compatible inputs and subject-disjoint probes where identifiers exist. CAUEEG uses recording-level folds and an annotated no-overlap sensitivity. Controls test stronger comparators, fully random initialization, label permutation, and scrambled-label adaptation. Fully-random and permutation controls are model- and task-targeted; adaptation controls are REVE-only.

\noindent\textbf{Main results.} On CAUEEG normal/mild cognitive impairment (MCI)/dementia classification ($N{=}\PaperValue{v5071b3c9d6a5451f}$ recordings), descriptive best-probe macro-area under the receiver operating characteristic curve (macro-AUROC) was $\PaperValue{vd29f27e84a2ae739}$--$\PaperValue{v5d9aa6cbb1c6f7dc}$ for five encoders versus $\PaperValue{vb8c001cb226923a9}$ for simple classical features ($\PaperValue{va4f8dff75a80ecd0}$ enhanced sensitivity). On the annotated no-overlap subset ($n{=}\PaperValue{v4cd043e2f21b8269}$), classical logistic regression remained above REVE ($\PaperValue{v91604c676610144e}$ vs.\ $\PaperValue{v105e6aae0134da16}$). Fully random REVE was above pretrained REVE in a separate repeated-split control ($\PaperValue{v420e51509322903f}$ vs.\ $\PaperValue{vf8e66e8a6ceffaac}$). Dataset-membership AUROC was $\PaperValue{vd0d193f424fcfe36}$ for every encoder in full space and after in-fold PCA-50; shuffled-label means were $\PaperValue{v82c48b60d09b5c24}$--$\PaperValue{vce6a66b1022fbfde}$ and balanced subsamples remained at $\PaperValue{v8c7c2b99622623a6}$. This demonstrates dataset membership, not causal site, geographic, or population effects. On CHB-MIT ($n{=}\PaperValue{v089d14b5891f734c}$), REVE reached $\PaperValue{vb370fc4c5a7faa29}$ versus enhanced GBM at $\PaperValue{v03ad6fa66a566705}$, fully random REVE at $\PaperValue{vea7ad4b657e4f2d6}$, and raw-signal random features at $\PaperValue{v0a71f280022bee49}$. The paired REVE--enhanced-GBM difference was $+\PaperValue{vfc5dd21abc2243b1}$~pp with 95\% CI $[-\PaperValue{v95de63f94c69e163},\PaperValue{ve3df06a4e857ec2a}]$~pp, which crosses zero. Per-epoch normalization also precluded an amplitude-aware comparator.

\noindent\textbf{Significance.} No task satisfies all four attribution conditions, so these results do not support a general claim of disease-information transfer across clinical populations. We distill the controls into a reporting protocol for clinical EEG foundation-model studies.

\smallskip\noindent\textbf{Keywords:} EEG, foundation models, transfer learning, clinical EEG, seizure detection, Alzheimer's disease, brain--computer interface, linear probing
\end{abstract}

\section{Introduction}

Electroencephalography (EEG) provides non-invasive, temporally resolved measurements of brain activity in clinical settings. A long-standing technical barrier to using EEG across studies is the difficulty of extracting features that transfer across participants and recording conditions~\cite{jayaram2016transfer}. This paper is about how claims of such transfer are evaluated, not about any particular clinical application: every task evaluated here is retrospective diagnostic classification on public or registration-gated cohorts, which is a precondition for downstream clinical use rather than a demonstration of it.

Foundation models are large neural networks pretrained on broad corpora, commonly with self-supervised objectives, to learn representations for downstream tasks. They have transformed natural language processing~\cite{bommasani2021opportunities} and computer vision~\cite{he2022mae}. In 2024--2025, several groups released pretrained EEG foundation models evaluated across brain--computer interface (BCI) and clinical downstream tasks~\cite{jiang2024labram,wang2025eegmamba,wang2025cbramod,ouahidi2025reve}.

Systematic multi-dataset comparisons now exist, but their protocols and adaptation settings differ~\cite{wu2025adabrain,kastrati2025eegbench,kontras2026neuroatlas}. We address a narrower gap: combining frozen probing, an external-cohort stress test, patient-overlap checks, and representation-level negative controls. Published evaluations report strong held-out-subject results on some clinical-monitoring tasks~\cite{yang2023biot,ouahidi2025reve}, but acquisition, population, and clinical composition can all change outside the benchmark cohort. AdaBrain-Bench standardizes comparisons but does not test this combination~\cite{wu2025adabrain}. Here, a \emph{targeted negative control} is an analysis designed to test one specific alternative explanation for an apparent model advantage, such as comparator weakness, architecture alone, label-independent structure, or nonspecific adaptation. Each control narrows one interpretation but neither excludes all confounding nor establishes causality; their joint pattern therefore supports only bounded attribution.

\subsection{Contributions}

In our review of the benchmark designs in Section~\ref{sec:related_work}, we did not identify one combining an external-clinical-cohort stress test with the present negative-control suite. We therefore test whether within-task frozen-probe advantages survive both. We evaluate LaBraM~\cite{jiang2024labram}, EEGMamba~\cite{wang2025eegmamba}, CBraMod~\cite{wang2025cbramod}, REVE~\cite{ouahidi2025reve}, and BIOT~\cite{yang2023biot} under task-specific frozen-probe protocols. Our contributions are:

\begin{enumerate}[noitemsep]

    \item \textbf{A targeted four-part control suite.} Stronger handcrafted comparators test baseline adequacy; matched fully random initialization tests pretrained-weight dependence; label permutation tests label dependence; and scrambled-label adaptation probes sensitivity to source-label correctness, with unmatched comparisons explicitly bounded. Coverage follows surviving task-specific claims rather than an exhaustive task-by-model grid.

    \item \textbf{A symmetric external-cohort stress test.} All five encoders and handcrafted families start from the same clean CAUEEG montage and use identical folds. Every encoder remains below the simple comparator, and the annotated no-overlap sensitivity preserves the classical-over-REVE ordering.

    \item \textbf{Dataset-identity controls.} Dataset membership is perfectly linearly decodable from every encoder after in-fold PCA-50, survives balanced subsampling, and collapses under label permutation. REVE-only Western--Western and harmonized-input extensions show that the effect is neither unique to the Western--Korean pair nor explained solely by line-frequency or absolute-scale cues; causal sources remain inseparable.

    \item \textbf{Bounded attribution of positive task margins.} On CHB-MIT, REVE shows pretrained-weight and label dependence, but its paired margin over enhanced GBM has an interval crossing zero. On ds004504, completed REVE and BIOT controls likewise leave positive AD-versus-FTD point margins unresolved. The suite therefore supports task- and pipeline-specific conclusions, not a general transfer claim.
\end{enumerate}

\begin{figure}[tbp]
\centering
\resizebox{0.925\textwidth}{!}{%
\begin{tikzpicture}[
  bx/.style={draw, rounded corners=1.4pt, align=center, font=\footnotesize, inner sep=3pt, line width=0.5pt},
  data/.style={bx, fill=cData, draw=cDataE, minimum width=4.2cm, minimum height=1.25cm},
  ext/.style={bx, fill=cExt, draw=cExtE, minimum width=4.2cm, minimum height=1.25cm},
  prep/.style={bx, fill=cPrep, draw=cRule, minimum height=1.0cm},
  chip/.style={bx, fill=cModel, draw=cModelE, minimum width=2.2cm, minimum height=0.58cm, font=\scriptsize},
  cchip/.style={bx, fill=cClass, draw=cClassE, minimum width=6.1cm, minimum height=0.58cm, font=\scriptsize},
  eval/.style={bx, fill=cEval, draw=cEvalE, minimum height=1.0cm},
  ctrl/.style={bx, fill=cCtrl, draw=cCtrlE, minimum width=6.5cm, minimum height=0.95cm},
  band/.style={font=\footnotesize\bfseries, anchor=west},
  blab/.style={font=\scriptsize\bfseries, text=black!70, anchor=west},
  sub/.style={font=\scriptsize, text=black!55, align=left},
  a/.style={-{Stealth[length=1.7mm]}, draw=cRule, line width=0.6pt},
]

\node[band] at (0,3.95) {Clinical EEG tasks};
\node[sub, anchor=west] at (3.85,3.95) {plus one auxiliary control dataset, which carries no task};

\node[data] (chb)  at (2.40,2.60) {\textbf{CHB-MIT}\\ictal vs.\ interictal\\{\scriptsize \PaperValue{v8c87773002afea40} ch $\cdot$ \PaperValue{v089d14b5891f734c} subj}};
\node[data] (tuab) at (7.15,2.60) {\textbf{TUH TUAB}\\normal vs.\ abnormal\\{\scriptsize \PaperValue{v6353b728a280369c} ch $\cdot$ \PaperValue{vbb379e938607be31} subj}};
\node[data] (ds)   at (11.90,2.60) {\textbf{ds004504}\\AD vs.\ HC / FTD\\{\scriptsize \PaperValue{v85f16b5dd393a698} ch $\cdot$ \PaperValue{v8b8e6dc8c36b6d62} / \PaperValue{v8b57fcf7d74c901f} subj}};
\node[data] (slp)  at (4.775,0.95) {\textbf{Sleep-EDF}\\wake vs.\ sleep\\{\scriptsize \PaperValue{v2335a20175692296} ch $\cdot$ \PaperValue{vd4d580014d7f8622} subj}};
\node[ext]  (cau)  at (9.525,0.95) {\textbf{CAUEEG}\\normal / MCI / dementia\\{\scriptsize \PaperValue{v9ad2a8acbb26ddba} ch $\cdot$ \PaperValue{v5071b3c9d6a5451f} rec.}};
\node[sub, anchor=west, text=cExtE] at (11.85,0.95) {external cohort};
\node[bx, fill=white, draw=cRule, dashed, minimum width=2.00cm, minimum height=1.25cm, align=center, font=\scriptsize] (aux) at (0.85,0.95) {\textbf{ds003490}\\identity control only\\{\scriptsize 19 ch $\cdot$ \PaperValue{v4b66988660e36874} subj}};

\node[prep, minimum width=14.3cm] (prep) at (7.15,-1.15) {\textbf{Task-specific preprocessing paths}\\\PaperValue{vd3860d9782b9539c}\,Hz and \PaperValue{v1d999751296ac196}\,s non-overlapping epochs; filtering, reference, rejection,\\and normalisation follow each dataset pipeline\\{\scriptsize exact operations and granularity: Section~\ref{sec:preproc}}};

\draw[a] (chb.south)  -- (chb  |- prep.north);
\draw[a] (tuab.south) -- ++(0,-0.55) -| (7.15,-0.35);
\draw[a] (ds.south)   -- (ds   |- prep.north);
\draw[a] (slp.south)  -- (slp  |- prep.north);
\draw[a] (cau.south)  -- (cau  |- prep.north);

\node[band] at (0,-2.60) {Representations};

\node[blab] at (0.55,-3.35) {Five frozen encoders};
\node[chip] (m1) at (1.80,-4.05) {LaBraM};
\node[chip] (m2) at (4.20,-4.05) {EEGMamba};
\node[chip] (m3) at (6.60,-4.05) {CBraMod};
\node[chip] (m4) at (3.00,-4.85) {REVE};
\node[chip] (m5) at (5.40,-4.85) {BIOT};
\node[sub, anchor=north west, text width=7.0cm] at (0.55,-5.28) {weights locked, task-specific embedding aggregation};

\node[blab] at (8.35,-3.35) {Simple plus targeted enhanced features};
\node[cchip] (k1) at (11.25,-4.05) {\textbf{simple}\quad band power $\cdot$ Hjorth $\cdot$ spectral entropy};
\node[cchip] (k2) at (11.25,-4.85) {\textbf{enhanced}\quad $+$ coherence $\cdot$ entropy $\cdot$ DFA $\cdot$ wavelet};
\node[sub, anchor=north west, text width=7.0cm] at (8.35,-5.28) {enhanced family excludes TUAB; nonlinear sensitivities reported separately};

\draw[a] (prep.south) -- ++(0,-0.45) -| (4.20,-3.73);
\draw[a] (prep.south) -- ++(0,-0.45) -| (13.40,-3.73);

\node[eval, minimum width=14.3cm] (probe) at (7.15,-6.85) {\textbf{Within-task split-aware probing}\\task-specific logistic settings; leave-one-subject-out or subject-grouped 5-fold where identifiers exist;\\CAUEEG recording-level with LDA/SVM-RBF and an annotated no-overlap sensitivity};

\draw[a] (4.20,-5.60) -- (4.20,-6.30);
\draw[a] (11.25,-5.60) -- (11.25,-6.30);

\node[band] at (0,-8.45) {Negative controls};
\node[sub, anchor=west] at (3.15,-8.45) {targeted, not exhaustive};

\node[ctrl] (c1) at (3.65,-9.85) {\textbf{Stronger comparators}\\{\scriptsize simple $\rightarrow$ enhanced}};
\node[ctrl] (c2) at (10.65,-9.85) {\textbf{Label permutation}\\{\scriptsize real $y$ $\leftrightarrow$ shuffled $y$}};
\node[ctrl] (c3) at (3.65,-11.40) {\textbf{Matched full random initialization}\\{\scriptsize pretrained vs.\ fully random}};
\node[ctrl] (c4) at (10.65,-11.40) {\textbf{Scrambled adaptation}\\{\scriptsize real vs.\ scrambled LoRA}};

\draw[a] (probe.south) -- (7.15,-8.95);
\draw[cRule, line width=0.6pt] (3.65,-8.95) -- (10.65,-8.95);
\draw[a] (3.65,-8.95)  -- (c1.north);
\draw[a] (10.65,-8.95) -- (c2.north);
\draw[a] (c1.south) -- (c3.north);
\draw[a] (c2.south) -- (c4.north);

\end{tikzpicture}%
}
\caption{Study design and control structure. Four benchmark datasets and Korean CAUEEG
follow documented task-specific preprocessing that is not uniform across tasks
(Section~\ref{sec:preproc}). Five frozen encoders and simple features cover every primary
task. The current enhanced family covers four benchmark tasks plus CAUEEG; TUAB is excluded
because no protocol-matched current-family result can be reconstructed. Independently,
TUAB's primary REVE and non-REVE rows use different samples. Probes and logistic settings are
task-specific; CAUEEG additionally reports LDA and SVM-RBF and uses recording-level folds
plus an annotated no-overlap sensitivity because general patient identifiers are unavailable.
Encoder coverage is incomplete: BIOT is unavailable on TUAB and Sleep-EDF. ds003490 has no
clinical task and enters only the Western--Western dataset-identity probe. Four targeted
controls challenge apparent advantages: stronger comparators across tasks, model-specific
label permutation and matched fully random initialisation (with a CHB-MIT raw-signal random-feature
reference), and REVE-only scrambled-label adaptation.}
\label{fig:design}
\end{figure}
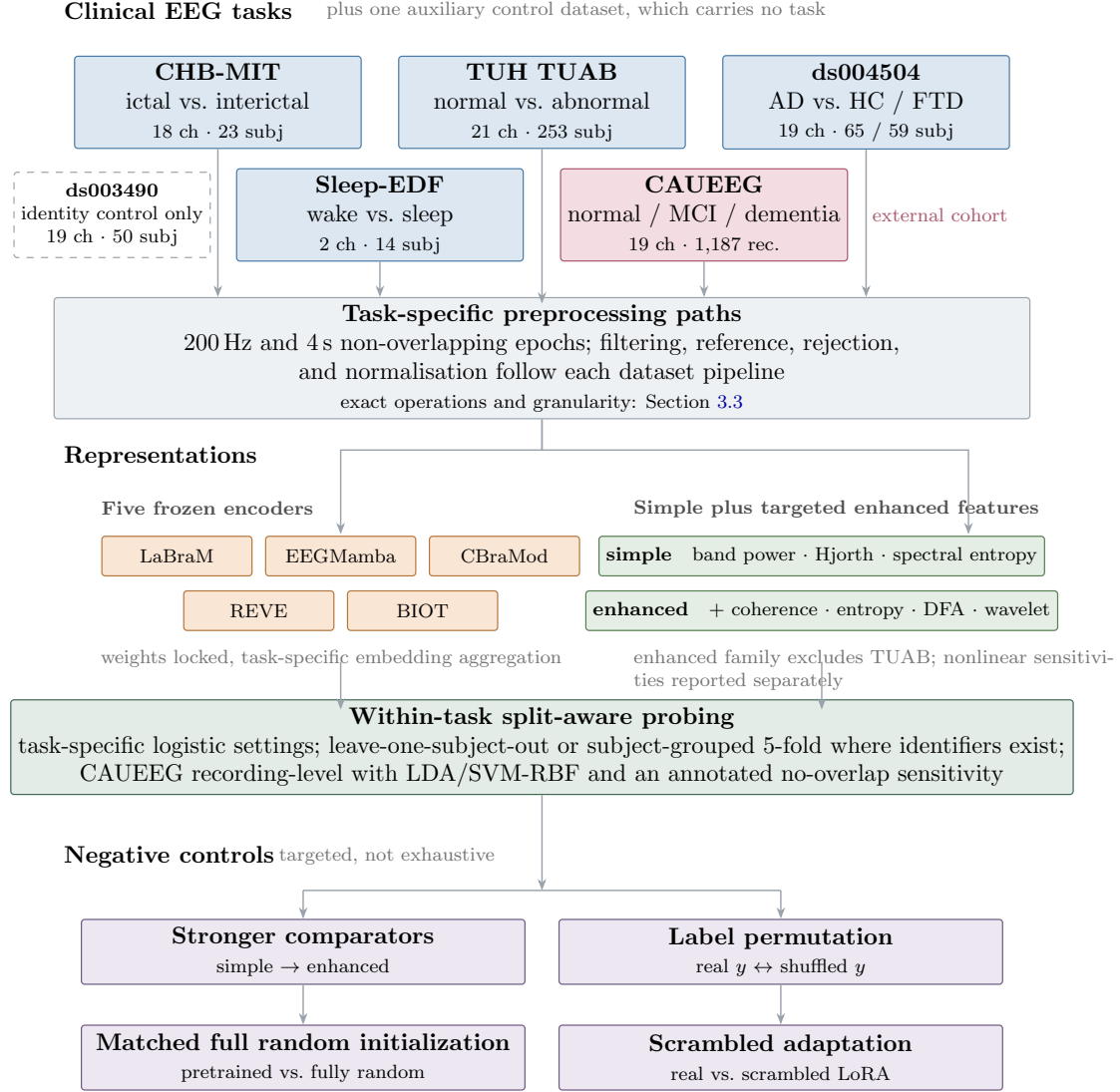

\noindent Figure~\ref{fig:design} summarises the study design: which representations are compared, under which splits, and which targeted controls interrogate apparent gains. The Methods section gives the full specification.

\section{Related Work}
\label{sec:related_work}

\subsection{EEG Foundation Models}

BIOT introduced a unified tokenizer for cross-dataset biosignal learning and was evaluated on EEG, ECG, and smartphone inertial-activity signals~\cite{yang2023biot}. LaBraM introduced vector-quantized spectral tokenization with masked prediction and reported results on abnormal and event classification, emotion recognition, and gait prediction~\cite{jiang2024labram}. CBraMod proposed criss-cross attention to model spatial and temporal dependencies separately~\cite{wang2025cbramod}; both LaBraM and CBraMod are evaluated in AdaBrain-Bench~\cite{wu2025adabrain}. EEGMamba used a bidirectional Mamba encoder with masked EEG reconstruction and reported results across six downstream tasks~\cite{wang2025eegmamba}; its backbone builds on selective state-space models~\cite{gu2024mamba}. REVE introduced 4D positional encoding designed to accommodate arbitrary electrode configurations and was pretrained on more than 60,000 hours from 92 datasets and approximately 25,000 subjects~\cite{ouahidi2025reve}.

\subsection{Evaluation Protocols}

Benchmark results vary across subjects, recording conditions, and datasets, and within-dataset evaluation does not by itself establish robustness to distribution shift, particularly when pretraining and evaluation share a source distribution~\cite{kastrati2025eegbench,kontras2026neuroatlas}. Leave-one-subject-out (LOSO) cross-validation keeps each test subject unseen during training and therefore directly evaluates cross-subject transfer. We treat within-dataset evaluation as complementary to, rather than a substitute for, external-cohort evaluation. AdaBrain-Bench established a standardized benchmark that includes clinical tasks (Temple University Hospital Abnormal EEG Corpus [TUAB] normal/abnormal, Siena seizure, and sleep staging) with cross-subject results for each~\cite{wu2025adabrain}. Its scope differs from ours in five respects: it evaluates four models (BIOT, EEGPT, LaBraM, CBraMod), not REVE or EEGMamba; it uses full fine-tuning as the primary adaptation strategy; its clinical scope covers anomaly detection, seizure detection, and sleep staging rather than neurodegenerative diagnosis; it does not include an external non-Western clinical dementia cohort comparable to CAUEEG; and it compares pretrained models with counterparts trained from scratch rather than testing matched pretrained and randomly initialized frozen encoders, label permutation, scrambled-label adaptation, or raw-signal random features~\cite{wu2025adabrain}. On clinical-monitoring tasks, AdaBrain-Bench found foundation models comparable to or worse than traditional models; on Sleep-EDF, its best foundation model reached 69.47\% versus 69.55\% for its best traditional model, directionally consistent with our 2-channel sleep result~\cite{wu2025adabrain}. AdaBrain-Bench also reported that linear probing generally underperformed full fine-tuning, with EEGPT exceptions on BCI-IV-2A and HMC, and concluded that current models were not ready for direct generalization through frozen representations~\cite{wu2025adabrain}.

\noindent\textbf{Benchmarks and critical evaluations of EEG foundation models.} Critical reviews find heterogeneous and limited evidence for off-the-shelf utility, with linear probes often weaker than fine-tuned models while scratch-trained specialists remain competitive~\cite{kuruppu2025review,liu2026progress}. Recent controlled benchmarks find only marginal advantages over supervised baselines such as EEGNet under causal, artifact-aware protocols~\cite{lee2025brainwave}, while properly tuned supervised models can match or beat EEG-FMs at a fraction of the parameters~\cite{wang2026audit}. Large multi-dataset comparisons report foundation models performing largely on par with simpler models, especially under clinical distribution shift~\cite{kastrati2025eegbench,kontras2026neuroatlas}. Other systematic benchmarks broaden task coverage and diagnostic analysis~\cite{xiong2025eegfmbench,lu2026omnieeg}, while robustness and interpretability audits show failure-mode-dependent behavior~\cite{sirca2026beyond}. Our study differs in three respects. \emph{(i)~An external non-Western stress test.} We evaluate pretrained encoders on Korean CAUEEG using a matched clean-input pipeline and add a fixed-split sensitivity on the annotated no-overlap held-out subset because general patient identifiers are unavailable for the recording-level primary analysis. \emph{(ii)~A targeted negative-control suite.} We combine stronger classical comparators with label permutation and task-specific scrambled-label, fully-random-initialization, and raw-signal random-feature controls. These test probe behavior, sensitivity to source-label correctness, and pretrained-weight dependence without treating either unmatched adaptation comparison as a causal estimate. \emph{(iii)~A dataset-identity probe.} Dataset membership remains linearly decodable at ceiling after PCA-50 for all five encoders on the Western--Korean pair and for REVE on the Western--Western pair. This is evidence for dataset identity, not a causal geographic, site, device, or population axis because those factors are confounded.

\noindent\textbf{Companion representation audit.} A companion preprint shares CAUEEG and some frozen-representation caches but asks a distinct spectral--temporal decoding question; it is therefore complementary rather than an independent replication~\cite{zare2026lrtc}.

\subsection{Linear Probing for Foundation Models}

Linear probing, training only a linear classifier on frozen pretrained features, is a standard evaluation protocol for foundation models in computer vision~\cite{chen2020simclr,he2022mae}. It assesses frozen features without task-specific encoder updates. For EEG, AdaBrain-Bench includes linear probing alongside full fine-tuning~\cite{wu2025adabrain}. We use frozen linear probing as the primary evaluation framework; explicitly labelled nonlinear and adaptation sensitivities are secondary.

\section{Methods}

Clinical EEG from four benchmark datasets is preprocessed through the documented task-specific pipelines, then evaluated with the applicable subset of five frozen pretrained encoders. Embeddings are evaluated by LOSO or subject-grouped five-fold probing when patient identifiers exist; CAUEEG uses recording-level folds and a separate fixed-split sensitivity on its annotated no-overlap held-out subset. Computation was performed locally on a Mac M3 laptop (16~GB unified memory), using CPU or MPS; EEGMamba's CUDA-only Mamba2 selective scan was reimplemented in pure PyTorch for this environment.

\noindent The Supplementary Material follows the order in which its contents are first used below.

\subsection{Foundation Models}

Five pretrained EEG foundation models with distinct architectures were evaluated on their applicable datasets within task-specific frozen-probe pipelines. Covered encoders use the same cohort, primary probe, and folds within each task except TUAB: REVE used a separate \PaperValue{vcbbc4d939e299c96}-epoch global subsample, whereas the classical, LaBraM, EEGMamba, and CBraMod rows used a \PaperValue{ve54ede0b638c1a02}-epoch subject-capped sample. TUAB model differences are therefore not treated as matched comparisons. BIOT is unavailable on TUAB because the retained input lacks the names required to construct its bipolar derivations, and on two-channel Sleep-EDF because the bipolar16 contract cannot be satisfied; all other model--task cells were evaluated. Task settings are specified in Section~\ref{sec:evaluation_protocol}. Fully-random-initialization and label-permutation controls target REVE on selected tasks and BIOT on both ds004504 contrasts; the raw-feature control is CHB-MIT-specific, and scrambled-label adaptation remains REVE-only.

\begin{table}[H]
\centering
\caption{Foundation models evaluated. Weights were obtained from publicly listed repositories; REVE required gated Hugging Face access. Parameter counts refer to the evaluated backbones as instantiated locally.}
\label{tab:models}
\footnotesize
\setlength{\tabcolsep}{2pt}
\begin{tabular}{@{}llccll@{}}
\toprule
\textbf{Model} & \textbf{Architecture} & \textbf{Params} & \textbf{Pretrain} & \textbf{Pretraining objective} & \textbf{Ref.} \\
\midrule
LaBraM-Base & BEiT-inspired Transformer & 5.8M & 2,500\,h & Masked token pred. & \cite{jiang2024labram} \\
EEGMamba & Bidir.\ Mamba2 & 3.3M & 16,724\,h & Masked reconst. & \cite{wang2025eegmamba} \\
CBraMod & Criss-Cross Attn. & 4.9M & 9,000+\,h & Masked reconst. & \cite{wang2025cbramod} \\
REVE-Base & 4D Pos Enc Transf. & 69.4M & $>$60,000\,h & Masked autoenc. & \cite{ouahidi2025reve} \\
BIOT & Bio Transformer & 3.2M & 6 EEG datasets$^\dagger$ & Contrastive + multitask & \cite{yang2023biot} \\
\bottomrule
\multicolumn{6}{@{}p{0.95\textwidth}@{}}{\footnotesize $^\dagger$The evaluated BIOT checkpoint was pretrained on MGH, SHHS, CHB-MIT, IIIC Seizure, TUAB, and TUEV. CHB-MIT and TUAB are therefore in-domain for BIOT.}
\end{tabular}
\end{table}

\noindent\textbf{LaBraM}~\cite{jiang2024labram} uses a vector-quantized spectral tokenizer with a masked-token transformer. We loaded checkpoint \texttt{braindecode/labram-pretrained} through Braindecode's LaBraM implementation (\url{https://braindecode.org/stable/generated/braindecode.models.LaBraM.html}, accessed 6 August 2026). The 129-entry spatial position tensor (128 channel positions plus one class-token position) is linearly interpolated to $N_{\mathrm{ch}}+1$ entries, and the 16-entry temporal position tensor is linearly interpolated to five positions for 4-s, \PaperValue{v9e88bdaf265f0667}-Hz inputs (\texttt{align\_corners=False}). These are study-specific approximations, not validated EEG channel- or time-mapping procedures.

\noindent\textbf{EEGMamba}~\cite{wang2025eegmamba} uses bidirectional Mamba2 state-space blocks~\cite{dao2024mamba2}. We use the pretrained masked-reconstruction model of Wang et al.\ (\emph{Neural Networks} 192:107816; Hugging Face \texttt{weighting666/EEGMamba}), not the end-to-end multi-task classifier of Gui et al.~\cite{gui2024eegmamba}. Its CUDA-only selective scan was reimplemented in pure PyTorch for CPU/MPS execution. The port uses channel-agnostic temporal and spectral projections with a convolutional positional encoding, so it accepts the 18-channel CHB-MIT montage and the \PaperValue{v85f16b5dd393a698}-channel montage; numerical parity with the upstream CUDA kernel was not independently tested.

\noindent\textbf{CBraMod}~\cite{wang2025cbramod} uses criss-cross attention (separate spatial and temporal attention), pretrained by masked reconstruction on more than 9{,}000 cleaned hours of the TUH EEG corpus. The evaluated public backbone instantiates 4.9M parameters in our local wrapper; Table~\ref{tab:models} reports this locally counted value.

\noindent\textbf{REVE}~\cite{ouahidi2025reve} uses 4D positional encoding (3D electrode coordinates plus temporal position) to accept varying electrode layouts without interpolation. REVE-Base (69.4M parameters, 512-d hidden states, 22 layers) was loaded via gated Hugging Face access. Extraction is task-specific: CHB-MIT mean-pools the final hidden states to 512 dimensions; the primary ds004504 analyses use the historical flattened $19\times4\times512=\PaperValue{v188e6eefc04fa197}$-dimensional output of the final whole-vector LayerNorm; and TUAB uses token-flattened embeddings. Consequently, dimensionality checks are interpreted only within the pipeline that produced them and are not transferred across tasks.

\noindent\textbf{BIOT}~\cite{yang2023biot} (3.2M parameters) is a biosignal transformer (linear attention, \PaperValue{v46852ea18dccb863}-d). We use its six-dataset EEG checkpoint, pretrained on MGH, SHHS, and the training sets of CHB-MIT, IIIC Seizure, TUAB, and TUEV. Its published \PaperValue{vc9b151715e9af58f}-token order is 16 longitudinal bipolar derivations followed by C3--A2 and C4--A1. We construct and order the first 16 by named subtraction, or reorder already-bipolar CHB-MIT inputs. A1/A2 are unavailable in the evaluated source montages, so the last two derivations are omitted without zero-padding; inputs without channel names are rejected. Because pretraining includes CHB-MIT and TUAB, those evaluations are in-domain. TUAB is not reported because the named-channel input required to construct the bipolar derivations is unavailable. We treat ds004504 and CAUEEG as out-of-domain subject to the published inventory's completeness.

\subsection{Datasets}

Table~\ref{tab:datasets} summarizes the benchmark, external-cohort, and auxiliary datasets and
their roles in the study.

\begin{table}[H]
\centering
\caption{Datasets. Benchmark datasets (top), the external CAUEEG stress-test cohort (middle), and one auxiliary dataset (bottom) that carries no clinical task and enters only the dataset-identity probe.}
\label{tab:datasets}

\small
\setlength{\tabcolsep}{3pt}

\begin{tabular*}{\textwidth}{@{\extracolsep{\fill}} >{\raggedright\arraybackslash}p{4.2cm} p{4.2cm} c c c >{\raggedright\arraybackslash}p{2.0cm}}
\toprule
\multicolumn{6}{l}{\textbf{Benchmark Datasets}} \\
\toprule
\textbf{Dataset} & \textbf{Task} & \textbf{Subj.} & \textbf{Ch.} & \textbf{Epochs} & \shortstack{\textbf{Access}/\\\textbf{origin}} \\
\midrule
CHB-MIT \cite{shoeb2010chbmit,physionet2010chbmit} & Ictal detection & \PaperValue{v089d14b5891f734c} & \PaperValue{v8c87773002afea40} & \PaperValue{v02e78d6b43b2c0cb} & PhysioNet \\
TUH TUAB \cite{obeid2016tuh,lopez2017tuab} & Normal vs.\ abnormal & \PaperValue{vbb379e938607be31} & \PaperValue{v6353b728a280369c} & \PaperValue{ve54ede0b638c1a02}/\PaperValue{vcbbc4d939e299c96}$^\dagger$ & TUH \\
ds004504 (OpenNeuro) \cite{miltiadous2023alzheimer} & AD vs.\ HC; AD vs.\ FTD & \PaperValue{v8b8e6dc8c36b6d62}/\PaperValue{v8b57fcf7d74c901f} & \PaperValue{v85f16b5dd393a698} & \PaperValue{v9309ff95441bded0}/\PaperValue{va341d39cd19b0dd6}$^\ddagger$ & OpenNeuro \\
Sleep-EDF \cite{kemp2000sleep} & Wake vs.\ sleep & \PaperValue{vd4d580014d7f8622} & \PaperValue{v2335a20175692296} & \PaperValue{v068a2220460adc9a}$^*$ & PhysioNet \\
\midrule
\multicolumn{6}{l}{\textbf{External-Cohort Stress Test}} \\
\midrule
CAUEEG \cite{kim2023caueeg} & Normal/MCI/Dementia & n/a$^\P$ & \PaperValue{v9ad2a8acbb26ddba}$^{\|}$ & \PaperValue{v5071b3c9d6a5451f}$^\P$ & South Korea \\
\midrule
\multicolumn{6}{l}{\textbf{Auxiliary Control Dataset (no clinical task)}} \\
\midrule
ds003490 (OpenNeuro) \cite{cavanagh2021ds003490} & Dataset-identity control only$^\S$ & \PaperValue{v4b66988660e36874} & 19 & --- & United States \\
\bottomrule
\multicolumn{6}{@{}p{\textwidth}@{}}{\footnotesize
$^\dagger$~From \PaperValue{vd01f79f27741aa43} available epochs, the classical, LaBraM, EEGMamba, and CBraMod rows use a \PaperValue{ve54ede0b638c1a02}-epoch subject-capped sample; REVE uses a separate \PaperValue{vcbbc4d939e299c96}-epoch global sample. $^*$~From 28,636 by a subject-balanced random cap of at most 214 epochs per retained subject. $^\ddagger$~AD/HC and AD/FTD use five-fold CV on \PaperValue{v9309ff95441bded0} and \PaperValue{va341d39cd19b0dd6} epochs, respectively. $^\P$~CAUEEG reports 1,155 patients in the full release but exposes no patient identifier; the analysed unit and tabulated count are recordings. $^{\|}$~CAUEEG exports 21 signals; all reported comparisons use the \PaperValue{v9ad2a8acbb26ddba}-channel 10--20 montage after excluding EKG and Photic (Section~\ref{sec:domain_generalization}). $^\S$~ds003490 supplies no diagnostic label, task, or metric anywhere in this paper; it appears only as the second member of the Western--Western pair in the dataset-identity probe (Section~\ref{sec:identity}).}
\end{tabular*}
\end{table}

\noindent\textbf{ds003490 (auxiliary control dataset)}~\cite{cavanagh2021ds003490}: This US resting-state and auditory-oddball cohort includes participants with Parkinson's disease and controls; we use 50 subjects at \PaperValue{v9ad2a8acbb26ddba} channels only for the REVE dataset-identity sensitivity (Section~\ref{sec:identity}), not for a clinical task. It was selected because it is absent from REVE's published pretraining inventory, whereas TDBRAIN is listed~\cite{ouahidi2025reve}. Each cohort retains its own \PaperValue{v9ad2a8acbb26ddba}-channel pipeline, so this Western--Western comparison tests whether dataset separability is unique to the Western--Korean pair, not a harmonised site or geographic effect; acquisition, preprocessing, population, and diagnostic composition remain confounded.

\noindent\textbf{What ``external cohort'' means here.}\label{sec:external_definition} Here, \emph{external cohort} denotes a separately constituted dataset not used in the five multi-dataset benchmark designs reviewed here~\cite{wu2025adabrain,kastrati2025eegbench,kontras2026neuroatlas,xiong2025eegfmbench,lu2026omnieeg}; it is not external validation of a fitted clinical model. Every primary clinical probe is trained and tested within one dataset. The dataset-identity probe crosses cohorts using dataset membership, not diagnosis, as its label. The Western-to-Korean LoRA analyses of Section~\ref{sec:lora} also cross cohorts, but both correct-versus-scrambled comparisons are unmatched descriptive sensitivities: the binary training schedules differ, and the three-way arms differ in retained seed provenance and downstream probe settings. CAUEEG is therefore a within-cohort stress test in a differently constituted population, not a clinical cross-cohort transfer evaluation.

\noindent\textbf{CHB-MIT}~\cite{shoeb2010chbmit,physionet2010chbmit}: Continuous scalp EEG from the full cohort of \PaperValue{v089d14b5891f734c} unique pediatric epilepsy patients at Boston Children's Hospital (all 24 released case directories; case chb21 was merged with chb01 because it is a later recording from the same patient, and chb12 was retained). The released recordings are sampled at \PaperValue{v46852ea18dccb863}\,Hz; our pipeline selects an \PaperValue{vc9b151715e9af58f}-channel double-banana bipolar montage, resamples it to \PaperValue{vfab46c67ff68f0b5}\,Hz, and applies a 60\,Hz notch. \textbf{Task: cross-subject ictal detection.} Positive epochs overlap an annotated seizure by $\geq$\PaperValue{v62a7533244504e8a}\%; negatives are interictal segments drawn from the \emph{same} seizure-containing recordings, at a 1:3 ictal:interictal ratio (per subject). All seizure-containing recordings per subject are included. Sharing the recording session reduces session confounding; $\pm30$\,s guard-band and all-files sensitivities further test that concern (Section~\ref{sec:chbmit_ictal}). CHB-MIT is absent from REVE's published pretraining inventory (the listed PhysioNet sources were Siena and ICARE~\cite{ouahidi2025reve}) and is treated as out-of-domain subject to the completeness of that inventory.

\noindent\textbf{CHB-MIT methods notes.} (i)~\emph{No amplitude-based rejection}: no amplitude-threshold rejection was applied because ictal epochs can be legitimately high-amplitude. The retained cohort contains \PaperValue{va0de40d901de51dc} ictal and \PaperValue{v1bf24db5d16e37da} interictal epochs, preserving the specified 1:3 ratio. (ii)~\emph{Amplitude normalization}: per-epoch z-scoring removes absolute amplitude from both encoder inputs and relative-band-power features. An amplitude-aware handcrafted comparator cannot be reconstructed from the stored normalized epochs and was not run; the comparison therefore does not establish superiority over such a baseline. (iii)~\emph{Bipolar coordinates}: double-banana derivations have no single-electrode coordinate. In the executed REVE pipeline, the \PaperValue{v8c87773002afea40} channels were assigned the first \PaperValue{v8c87773002afea40} entries of its standard 10--20 coordinate bank in index order rather than derivation midpoints. The assigned coordinate matches one constituent electrode for 6 of 18 derivations (the first electrode for 4) and neither electrode for 12. This task therefore does not test REVE under anatomically valid bipolar coordinates, and no alternate-coordinate sensitivity was run.

\textbf{Label note:} All CHB-MIT subjects are epilepsy patients (pediatric intractable-epilepsy cases recorded during pre-surgical evaluation with anticonvulsant withdrawal); the dataset contains \emph{no} non-epileptic controls~\cite{shoeb2010chbmit}. Positive epochs are \emph{ictal}: they overlap an annotated seizure interval by $\geq$50\%; negative epochs are interictal segments from the \emph{same} seizure-containing recordings. The task is therefore \textbf{cross-subject ictal detection} (ictal vs.\ same-session interictal epochs, in held-out epilepsy patients), not detection of a persistent epileptic-brain trait and not sample-precise onset localization from short windows, which is a harder, separately defined task.

\noindent\textbf{TUH Abnormal (TUAB)}~\cite{obeid2016tuh,lopez2017tuab}: Clinical hospital EEG from Temple University Hospital. Our retained evaluation cohort contains \PaperValue{vbb379e938607be31} subjects and \PaperValue{v6353b728a280369c} channels from the 10--20 system. Binary normal/abnormal labels are derived from clinical EEG reports and assessments~\cite{lopez2017tuab}. This dataset shares clinical population and recording protocol with the TUEG corpus used for CBraMod, LaBraM, and EEGMamba pretraining~\cite{jiang2024labram,wang2025cbramod,wang2025eegmamba}. Cross-referencing against the published pretraining dataset list~\cite{ouahidi2025reve}, TUH is confirmed as the largest source in REVE's pretraining corpus (14,987 subjects, 26,847 hours, 44\% of all pretraining data). The REVE authors removed downstream evaluation data from pretraining, but distributional similarity remains. BIOT was also pretrained on TUAB~\cite{yang2023biot}. TUAB is therefore in-domain or distributionally adjacent for all five models; results should not be compared directly with confirmed out-of-domain evaluations.

\noindent\textbf{OpenNeuro ds004504}~\cite{miltiadous2023alzheimer}: Eyes-closed resting-state EEG recorded with \PaperValue{v9ad2a8acbb26ddba} channels in the standard 10--20 montage. The dataset comprises \PaperValue{v79692b8e09fc9f85} participants with Alzheimer's disease, \PaperValue{v089d14b5891f734c} with frontotemporal dementia (FTD), and \PaperValue{v65e94ef249bf898d} healthy controls. The binary AD-vs-HC task uses $n{=}65$ and \PaperValue{v9309ff95441bded0} epochs; the AD-vs-FTD task uses $n{=}\PaperValue{v8b57fcf7d74c901f}$ and \PaperValue{va341d39cd19b0dd6} epochs; and the three-way AD/FTD/HC control uses all $n{=}88$ participants. Recordings were ICA-preprocessed. Both binary tasks use five-fold stratified group cross-validation with zero subject overlap between folds. Because ds004504 is absent from the published REVE pretraining inventory (cross-referenced against all 56 OpenNeuro datasets in~\cite{ouahidi2025reve}), this evaluation does not involve direct dataset overlap; however, the present comparison does not establish that pretraining caused the observed separability.

\noindent\textbf{Sleep-EDF}~\cite{kemp2000sleep}: Whole-night polysomnographic recordings from 20 healthy subjects. The analysis uses a fixed \PaperValue{v86bbfd61d5d0eb68}-subject cohort from the processed input available when the primary task was constructed; every retained subject has at least six epochs of both wake and sleep. A subsequently available fifteenth eligible record was not added to this fixed comparison set. Two EEG channels (Fpz-Cz, Pz-Oz) are used. Original 30\,s annotations (W, N1, N2, N3, REM) were binarized to Wake vs.\ Sleep, then re-epoched into \PaperValue{v57817cece7ef7fe2}\,s windows with majority-vote labeling. \textbf{Four task-construction choices materially shape the result.} (i)~Only the first recording night per subject is used. (ii)~Surrounding wake is cropped to 30 minutes on either side of the sleep period, so the task does not represent natural nocturnal prevalence. (iii)~When wake exceeds sleep by more than 3:1, wake is subsampled to exactly 3:1 using seed 42. (iv)~One advancing \texttt{RandomState(42)} samples at most $\lfloor\PaperValue{vade98305d10334de}/\PaperValue{v86bbfd61d5d0eb68}\rfloor=214$ epochs from each subject, yielding \PaperValue{va9332b36af636f2b} epochs; this cap is subject-balanced, not label-stratified. Normalisation is per channel across all accepted epochs of one subject. These choices define a cropped, class-capped, subject-capped two-channel task rather than deployment-prevalence sleep classification.

\noindent\textbf{CAUEEG (Chung-Ang University Hospital EEG)}~\cite{kim2023caueeg}: Clinical EEG database from Chung-Ang University Hospital, Seoul, South Korea, comprising 1,379 recordings from 1,155 patients. The three-way task retains \PaperValue{v5071b3c9d6a5451f} recordings: 459 Normal, 417 mild cognitive impairment (MCI), and 311 Dementia. The 192 excluded recordings carry primary categories outside that task (93 Parkinsonian syndrome, 72 transient global amnesia, \PaperValue{v86bbfd61d5d0eb68} frontotemporal dementia, and 13 normal-pressure hydrocephalus). The public export includes \PaperValue{v6353b728a280369c} signals at \PaperValue{vfab46c67ff68f0b5}\,Hz; our clean comparison retains \PaperValue{v9ad2a8acbb26ddba} standard 10--20 EEG channels and excludes EKG and Photic. CAUEEG exposes no patient identifier, so the main comparison is recording-level; a fixed-split sensitivity uses the public no-overlap annotation (Section~\ref{sec:domain_generalization}). CAUEEG is absent from the enumerated pretraining inventories of all five encoders, subject to their completeness~\cite{jiang2024labram,wang2025eegmamba,wang2025cbramod,ouahidi2025reve,yang2023biot}.

\subsection{Preprocessing}
\label{sec:preproc}

Preprocessing was task-specific; no single full chain was applied to all five datasets. All primary pipelines used MNE-Python~\cite{gramfort2013mne}, 200\,Hz signals, and 4.0\,s non-overlapping epochs. CHB-MIT used 0.5--70\,Hz filtering, a 60\,Hz notch, common-average reference (CAR), no amplitude-threshold rejection, and per-epoch, per-channel z-scoring followed by clipping at $\pm8$. TUAB was processed using the executed 0.5--70\,Hz/CAR pipeline without a notch filter. It also used no amplitude-threshold rejection; z-scoring and clipping were performed per channel within each recording. The ds004504 path began from the released ICA-preprocessed derivatives, then used 0.5--70\,Hz filtering, a \PaperValue{v4033d5c509342cb0}\,Hz notch, CAR, and $500\,\mu$V peak-to-peak rejection before per-recording z-scoring and clipping. CAUEEG used its native 200\,Hz sampling, a 0.5\,Hz high-pass and 60\,Hz notch on the clean \PaperValue{v85f16b5dd393a698}-channel montage, with no 70\,Hz low-pass, CAR, amplitude rejection, z-scoring, or clipping.

Sleep-EDF used a separate pipeline: resampling to 200\,Hz, 0.5--70\,Hz bandpass filtering, no notch filter, no CAR, and no amplitude-threshold rejection, followed by per-channel z-scoring across all accepted epochs of one subject and clipping at $\pm8$. Sleep-EDF retains the historical unnotched input because the primary analysis preceded the later 50\,Hz-notch revision and subsequent comparator rows were matched to that version. No notch-filter sensitivity was performed for either task; their absolute estimates may therefore retain mains-frequency information.\footnote{These are the study's downstream pipelines; they do not reproduce every encoder's pretraining preprocessing. Here ``raw-signal'' denotes the preprocessed time-series supplied directly to the control rather than an encoder, not unprocessed ADC output.}

For CAUEEG, each recording was divided into all available non-overlapping 4-s epochs. Epoch-level encoder embeddings and handcrafted feature vectors were averaged within the recording before fold assignment, yielding one vector and one diagnosis label per recording. For the clinical matrix, every representation was standardized within each training fold; REVE alone was additionally reduced by in-fold PCA-50 because its recording vector is 38,912-dimensional. The other encoder and handcrafted vectors entered the probes without PCA.

\subsection{Classical Baseline}

To compare predictive performance with foundation-model representations, we use handcrafted features: relative band power in $\delta$ (0.5--4\,Hz), $\theta$ (4--8\,Hz), $\alpha$ (8--13\,Hz), $\beta$ (13--30\,Hz), and $\gamma$ (30--45\,Hz) computed by Welch's method~\cite{welch1967use}; Hjorth parameters (activity, mobility, complexity)~\cite{hjorth1970eeg}; Shannon spectral entropy~\cite{inouye1991spectral}; and two study-defined, channel-index-based summaries (an alpha-asymmetry proxy and mean theta power over the first four channels). The feature family has $9 \times N_\text{ch}+2$ features per epoch (\PaperValue{v2c6439cdc1da1ac5} features for a \PaperValue{v9ad2a8acbb26ddba}-channel montage). For Sleep-EDF, the task-specific baseline uses the $9\times N_\text{ch}=\PaperValue{vc9b151715e9af58f}$ spectral, Hjorth, and entropy features without the two study-defined summaries. Exact formulas, channel rules, spectral settings, and enhanced-feature parameters are specified in Supplementary Material, Section~S1 (``Classical-feature and CAUEEG probe specifications''). We use L2-regularized logistic regression with balanced class weights.

To test whether more sophisticated handcrafted features could close the gap with foundation models, we also evaluate an \textbf{enhanced classical baseline} adding magnitude-squared coherence summaries over at most 20 channel pairs~\cite{welch1967use}, sample entropy~\cite{richman2000sample}, an epoch-scale detrended fluctuation analysis (DFA) summary~\cite{peng1995dfa}, and study-defined Haar-basis wavelet energy~\cite{mallat1989wavelet}. The full family has $15 \times N_\text{ch} + 6$ features per epoch. The four-second DFA statistic is retained as a short-window summary and is not interpreted as evidence of long-range temporal scaling. For the matched Sleep-EDF comparison, the 16 advanced summaries are appended to its \PaperValue{vc9b151715e9af58f}-feature task-specific baseline, yielding \PaperValue{vd9365d4b75df7fc6} features; a separate \PaperValue{vf40da55154906991}-versus-\PaperValue{v79692b8e09fc9f85}-feature sensitivity evaluates the two study-defined summaries in both families. Each comparison uses the same logistic classifier as its task-specific simple baseline.

\subsection{Evaluation Protocol}\label{sec:evaluation_protocol}

\noindent \textbf{Frozen probing.} Each pretrained encoder is frozen (no gradient updates). The primary probe is L2-regularized logistic regression (\texttt{class\_weight=balanced}, \texttt{solver=lbfgs}, \texttt{max\_iter=1000}): $C\,=\,\PaperValue{v943ac31c08c49693}$ for CHB-MIT and Sleep-EDF and $C\,=\,0.1$ for TUAB, the notch-filtered ds004504 tasks, and CAUEEG. These pipeline defaults were not selected by nested cross-validation, a limitation when comparing absolute estimates across tasks. The CHB-MIT result is insensitive to $C$ over $\{0.1,\PaperValue{v943ac31c08c49693},10\}$ (mean-fold AUROC $\PaperValue{vb370fc4c5a7faa29}$). CAUEEG sensitivities use fold-local StandardScaler followed by LDA (\texttt{solver=lsqr}, automatic shrinkage) or SVM-RBF ($C{=}10$, $\gamma=\texttt{scale}$, balanced class weights, probability estimates, seed 42); any REVE PCA is fitted within the same training fold. Best-of-family comparisons and nonlinear classical heads are descriptive rather than a unified linear-probe estimator. Balanced class weights are used by the logistic and CAUEEG SVM probes; the AD-versus-HC GradientBoosting sensitivity uses balanced sample weights, whereas the CHB-MIT GradientBoosting and MLP sensitivities are unweighted. CHB-MIT first samples same-session negatives to a fixed 1:3 ictal:interictal ratio; no additional under- or oversampling is applied during probing.

\noindent  \textbf{Metrics.} Balanced accuracy (BA) and AUROC are reported as mean $\pm$ standard deviation across folds. The analysis unit differs by task and is specified below. All five benchmark tasks are evaluated at the epoch level within subject-disjoint folds. CAUEEG diagnosis is evaluated per recording; dataset identity uses one unit per ds004504 subject and one per CAUEEG recording (\PaperValue{v8cc1fbf12d281798} and \PaperValue{v5071b3c9d6a5451f} units, respectively). For CHB-MIT and Sleep-EDF, every subject contributes both classes, so the reported estimand is epoch-level discrimination in held-out subjects; no single subject-level outcome was defined. Where aggregation is defined, subject- and epoch-level estimands are not interchangeable: across the two ds004504 contrasts, pooling epoch probabilities raises AUROC for both models, while the between-model ordering reverses only on AD versus HC (Sections~\ref{sec:ad_band_ablation} and~\ref{sec:limitations}).

The prediction unit follows the task construction and available metadata. The five benchmark tasks use short epochs as prediction units because the executed frozen-representation pipelines operate on short windows; folds are subject-disjoint, so no participant contributes epochs to both training and testing. For CHB-MIT and Sleep-EDF, each participant contributes both classes, so a single subject-level outcome is not defined. TUAB and ds004504 epochs inherit recording- or subject-level diagnoses; their primary estimates therefore quantify segment-level discrimination in held-out subjects rather than subject-level diagnostic performance. Subject-aggregated estimates are reported separately for the two ds004504 contrasts, where retained predictions support that estimand. CAUEEG is evaluated per recording because diagnoses are released at the recording level and general patient identifiers are unavailable; epoch representations are averaged within each recording before splitting.

Comparisons are interpreted within tasks, where representations share the outcome, analysis unit, and validation design, subject to the stated channel contracts and the TUAB sample mismatch. AUROCs are not compared across tasks because the outcomes, populations, analysis units, and validation schemes differ; epoch-, recording-, and subject-level values are distinct estimands and are not pooled into an overall model ranking. Cohort and epoch counts are in Table~\ref{tab:datasets}, and preprocessing is specified in Section~\ref{sec:preproc}.

\noindent\textbf{CHB-MIT.} The primary analysis uses epoch-level patient LOSO. REVE tokens are
mean-pooled to 512 dimensions, and every encoder representation is reduced by in-fold PCA-200
before probing. Interictal negatives come from the seizure-containing recordings at the fixed 1:3 ratio stated above.

\noindent\textbf{TUAB.} The analysis uses epoch-level, subject-grouped five-fold validation and
token-flattened REVE embeddings. REVE uses \PaperValue{vcbbc4d939e299c96} epochs, whereas the classical, LaBraM,
EEGMamba, and CBraMod rows use \PaperValue{ve54ede0b638c1a02}; model differences are therefore not estimated.

\noindent\textbf{ds004504.} Both AD-versus-HC and AD-versus-FTD use epoch-level, subject-grouped
five-fold validation and the historical \PaperValue{v188e6eefc04fa197}-dimensional flat-LayerNorm REVE representation. The primary estimand is
epoch-level discrimination; subject-aggregated estimates are reported separately where retained predictions permit it.

\noindent\textbf{Sleep-EDF.} Wake-versus-sleep uses epoch-level subject LOSO and retains epoch
vectors without an additional reduction. The first-night, two-channel construction and its wake,
label-ratio, and per-subject caps are specified in the Sleep-EDF dataset description above.

\noindent\textbf{CAUEEG.} The primary analysis uses recording-level stratified five-fold validation,
with identical folds across representations. Epoch vectors are averaged within recording; REVE alone
uses the fold-local dimensionality reduction specified in Section~\ref{sec:preproc}. The fixed-split patient-overlap sensitivity is reported separately.

Together, these protocols keep training and test subjects disjoint wherever identifiers exist, while
retaining the analysis unit supported by each task's labels and metadata. They do not create a common
cross-task estimand, and the TUAB sample mismatch remains an explicit exception to within-task matching.

%
%

\subsection{The Negative-Control Suite}
\label{sec:controls_methods}

A benchmark number can be high for reasons other than pretrained representation quality: a weak
comparator, the architecture alone, probe capacity on high-dimensional features, or an adaptation
procedure that gives similar target performance after source-label permutation. Each control below
targets one of these explanations within its stated coverage; Section~\ref{sec:controls} reports the
corresponding results.

Each control addresses a distinct alternative explanation and is interpreted only within its stated
coverage. Controls are allocated sequentially to active positive model-specific claims; once a claim fails an earlier comparator, uncertainty, or provenance gate, later controls are not required to reject that claim.

\noindent\textbf{Comparator adequacy.} Stronger handcrafted features test whether an apparent encoder
advantage reflects a weak baseline. The current enhanced handcrafted family covers four of the five benchmark tasks and
the CAUEEG stress test; TUAB is not included because a protocol-matched current-family result could not be reconstructed from the retained data. CAUEEG additionally includes an SVM-RBF nonlinear-probe sensitivity, whereas
Gradient Boosting and MLP sensitivities were run only on CHB-MIT and AD versus HC (Section~\ref{sec:enhanced}).

\noindent\textbf{Pretrained-weight dependence.} Matched fully random initialization tests whether pretrained
weights outperform the same architecture under matched inputs, folds, and probes. REVE was tested on
CHB-MIT, CAUEEG, and AD versus FTD; BIOT was tested on AD versus HC and AD versus FTD (Section~\ref{sec:randominit}).

\noindent\textbf{Label dependence.} Label permutation tests whether probe performance depends on the
observed label structure rather than representation dimensionality or probe capacity. Labels were
shuffled within subject for CHB-MIT and the complete PCA-200--scaling--logistic pipeline was refitted inside each fixed LOSO fold. For ds004504, BIOT was tested in the primary epoch-level pipelines for AD versus HC and AD versus FTD by permuting one diagnosis per participant, broadcasting it to that participant's epochs, and reusing the fixed folds. REVE uses direct \PaperValue{v188e6eefc04fa197}-dimensional subject-mean controls for AD versus HC, binary AD versus FTD, and three-way AD/FTD/HC. No Control~3 result is reported for TUAB, Sleep-EDF, or CAUEEG (Section~\ref{sec:labelperm}).

\noindent\textbf{Adaptation-label dependence.} Scrambled-label adaptation tests whether adaptation
depends on correct source labels by repeating adaptation after source-label permutation. This
REVE-only control was evaluated on binary and three-way CAUEEG, but neither completed comparison is a one-change label-effect experiment: the binary training schedules differ, and the three-way comparison differs in retained seed provenance and downstream SVM settings (Section~\ref{sec:lora}).

\noindent\textbf{Interpretation and analysis units.} These controls follow the model-specific claims that survive earlier attribution gates. Comparator and fully-random-initialization contrasts are descriptive unless a paired interval is stated; label permutation provides a finite-randomization result under its stated probe; and the adaptation aggregates are unpaired and are not equivalence tests. The ds004504 BIOT controls and REVE random-initialization comparison follow the primary epoch-level folds; the REVE label-permutation sensitivities aggregate its primary flat-LayerNorm vectors to one vector per participant. CAUEEG controls are recording-level. The CHB-MIT Gaussian raw-feature reference does not bound other raw-signal representations.


\noindent\textbf{Interpretive attribution framework.}\label{sec:attribution_rubric}
We apply four conditions in sequence to each positive model-specific margin: (i)~the encoder exceeds the stronger tested handcrafted family under the task's specified probe and folds; (ii)~paired uncertainty supports a positive margin where the analysis unit permits; (iii)~the published pretraining inventory shows no overlap with the evaluation cohort or a closely related source distribution; and (iv)~the pretrained representation exceeds matched random initialization and its observed-label probe exceeds the corresponding permutation null, with a matched scrambled-label comparison for adaptation claims. Failure at an earlier gate ends the positive attribution claim; later controls can still characterize the pipeline but cannot rescue that claim. The framework was formulated after the analyses and is not a prospective success criterion or an AUROC threshold. Section~\ref{sec:success_criterion} summarizes the model-specific outcomes.

\section{Results}

This section is organised by \emph{control}, not by dataset. Section~\ref{sec:benchmark_estimates}
reports the benchmark estimates once, in a single table, because those estimates are the object the
rest of the section interrogates rather than the finding itself. Section~\ref{sec:controls} then
applies the four negative controls defined in Section~\ref{sec:controls_methods}, and
Sections~\ref{sec:domain_generalization}--\ref{sec:overlap} report the external-cohort stress test
and the two confound probes. Additional balanced-accuracy results are consolidated in
Supplementary Material, Section~S2 (``Additional balanced-accuracy results''). Parameter counts are
reported in Table~\ref{tab:models}, AUROC estimates in Table~\ref{tab:benchmark}, and analysis units,
validation splits, and reporting rules in Section~\ref{sec:evaluation_protocol}.

\subsection{Benchmark Estimates}
\label{sec:benchmark_estimates}

Table~\ref{tab:benchmark} consolidates the frozen-probe estimates that the subsequent
controls interrogate.

\begin{table}[H]
\centering
\caption{Frozen-probe estimates for all six primary tasks and every evaluated representation. Entries are comparable within a task except on TUAB, where REVE and the other rows use different epoch samples; no overall ranking is computed.}
\label{tab:benchmark}
\footnotesize
\setlength{\tabcolsep}{4pt}
\resizebox{\textwidth}{!}{%
\begin{tabular}{@{}lcccccc@{}}
\toprule
& \multicolumn{6}{c}{\textbf{AUROC (\%), mean $\pm$ fold SD}} \\
\cmidrule(l){2-7}
\textbf{Task (dataset)} & \textbf{Classical} & \textbf{LaBraM} & \textbf{EEGMamba} & \textbf{CBraMod} & \textbf{BIOT} & \textbf{REVE} \\
\midrule
Ictal vs.\ interictal (CHB-MIT) & $\PaperValue{v353db98557019530} \pm \PaperValue{v3db52c932664da95}$ & $\PaperValue{v729fba494eb3b3a0} \pm \PaperValue{v7d45acbdc13c4d10}$ & $\PaperValue{vc1d63144fdb94693} \pm \PaperValue{ve9a03c4b0ed0ec4e}$ & $\PaperValue{v6e48c0916b1d2d4f} \pm \PaperValue{v733f49bd131c1794}$ & $\PaperValue{vd96fe4f01f5247e1} \pm \PaperValue{v81ccfb305f7340cd}^{\ast}$ & $\PaperValue{vbf3bbb54944f7133} \pm \PaperValue{vf3452f892263c4e7}$ \\
Normal vs.\ abnormal (TUAB) & $\PaperValue{v7f7936b17cbdaa1f} \pm \PaperValue{v5c7bfe3bf037c31a}$ & $\PaperValue{vbc319544810195ec} \pm \PaperValue{v69e9d2d7bfd9fa79}^{\ast}$ & $\PaperValue{v62252131128de21a} \pm \PaperValue{v779c63b5642ab262}^{\ast}$ & $\PaperValue{v5c716615cad2df62} \pm \PaperValue{v5c152b4002418033}^{\ast}$ & --- & $\PaperValue{v2c91484c325db6c8} \pm \PaperValue{v361a7d67b6215036}^{\ast}$ \\
AD vs.\ HC (ds004504) & $\PaperValue{ve8907089e5fdcc61} \pm \PaperValue{v96559ce4c32d22e8}$ & $\PaperValue{v63effdd9a593ad1d} \pm \PaperValue{vc79915caa9a7bd27}^{\dagger}$ & $\PaperValue{v9bfcf9c07ea7bc8e} \pm \PaperValue{v32bed4db5f25c7b7}$ & $\PaperValue{v58081b13c1e8f7fb} \pm \PaperValue{va1c49c84f59c32c1}$ & $\PaperValue{vb19d0558443f5600} \pm \PaperValue{v863bbc5986a123e3}$ & $\PaperValue{v6c4d910ddadcf793} \pm \PaperValue{v67f226f646217c2e}$ \\
AD vs.\ FTD (ds004504)$^{\P}$ & $\PaperValue{v5bef026461f3d681} \pm \PaperValue{v226893c4a2e2fc65}$ & $\PaperValue{v2f99e2ff8b4fe338} \pm \PaperValue{v57ca568f41ac64da}$ & $\PaperValue{vc76775337296598a} \pm \PaperValue{v9dd56a66a888b39f}$ & $\PaperValue{v88ff10ac5fe560fa} \pm \PaperValue{vd4e1346b00fccff9}$ & $\PaperValue{v936f50c3c6559f35} \pm \PaperValue{v56b1335444acc45a}$ & $\PaperValue{v9c0f67aae64c8b39} \pm \PaperValue{v006a6bb0055b1c0b}$ \\
Wake vs.\ sleep (Sleep-EDF) & $\PaperValue{v780df097635c7608} \pm \PaperValue{vc76e0557a3b9825e}$ & $\PaperValue{v1554fd9059e50c59} \pm \PaperValue{v802caa2e93753a12}$ & $\PaperValue{v6eba6c716a0e2e69} \pm \PaperValue{v717677d42f14735f}$ & $\PaperValue{v94496782c36c58cc} \pm \PaperValue{v0ed9f62fb1f14c3a}$ & --- & $\PaperValue{ve57fc765a91cb8dc} \pm \PaperValue{ve80637b0015c0e39}$ \\
N/MCI/dementia (CAUEEG)$^{\ddagger}$ & $\PaperValue{vc6ca7300571d5b6e} \pm \PaperValue{va1268b7d9f38d137}$ & $\PaperValue{ve8bc7acec10e28af} \pm \PaperValue{v8724eed53aa08f27}$ & $\PaperValue{v7a1ed695d9e055b1} \pm \PaperValue{v05dcb455ceb71af7}$ & $\PaperValue{vda7643566ef0f63c} \pm \PaperValue{ve80dfe151b5c24b2}$ & $\PaperValue{ve7e6246d845f1eb4} \pm \PaperValue{vd5b9323303ec3f3f}$ & $\PaperValue{v429f8cb22b105cdb} \pm \PaperValue{vada72137c6342650}$ \\
\bottomrule
\end{tabular}}

\vspace{3pt}
{\scriptsize\raggedright $^{\ast}$Confirmed pretraining overlap with the evaluation cohort; not out-of-domain transfer. TUAB REVE uses a separate \PaperValue{vcbbc4d939e299c96}-epoch global sample; the classical, LaBraM, EEGMamba, and CBraMod cells use a \PaperValue{ve54ede0b638c1a02}-epoch subject-capped sample, so TUAB cells are not directly differenceable. $^{\dagger}$Non-deterministic on Apple MPS; centre of a $[\PaperValue{vc0d17f37b42efdb9},\,\PaperValue{v675121b0099df864}]$ spread, excluded from every ranking claim. $^{\ddagger}$External cohort, recording-level folds, best of LogReg/LDA/SVM-RBF per family; Supplementary Material, Section~S3.1 (``Complete clean-input probe matrix'') reports the full matrix. \hbox{---}~not run because BIOT's input contract cannot be met on that task. Classical is the simple handcrafted family; the enhanced family and nonlinear heads are in Section~\ref{sec:enhanced}. Controls (Section~\ref{sec:controls}) were applied to task-specific REVE and BIOT claims. $^{\P}$The AD-vs-FTD row uses \PaperValue{va341d39cd19b0dd6} epochs from \PaperValue{v8b57fcf7d74c901f} subjects.\par}
\end{table}

\noindent\textbf{Pretraining provenance.} Two rows carry exposure that bars an out-of-domain reading. CHB-MIT was included in BIOT's pretraining~\cite{yang2023biot}, so its top rank on that task is not evidence of transfer. On TUAB the overlap is universal: LaBraM, CBraMod, and EEGMamba were pretrained on TUEG~\cite{jiang2024labram,wang2025cbramod,wang2025eegmamba}, REVE's corpus includes TUH (14{,}987 subjects, 44\% of corpus)~\cite{ouahidi2025reve}, and BIOT was pretrained on TUH~\cite{yang2023biot}. Every TUAB entry is therefore provenance-limited, and none should be compared against a confirmed out-of-domain estimate. The LaBraM AD-vs-HC entry carries a separate reproducibility caveat: its 129-entry position tensor is interpolated to 20 entries with \texttt{F.interpolate}, which is non-deterministic on Apple MPS, and five load-and-extract cycles on identical input with a fixed CV seed give $\PaperValue{v392bd82ba0d864e3} \pm \PaperValue{va77236db5a776f0c}$.

\noindent\textbf{Reading the benchmark and subsequent controls.} Four observations orient the control
analyses, and none is an overall ranking. First, the largest directly comparable descriptive exceedances of the simple classical column occur on
CHB-MIT and the AD-versus-FTD contrast. On AD versus HC, CBraMod,
BIOT-bipolar16, and REVE are also \PaperValue{vbfa278f28ef1e2b9}--\PaperValue{v0492ae6fcff07eb5}~pp above simple classical, but the enhanced
classical family reaches \PaperValue{v45a4ef01b74d8208}\% (Table~\ref{tab:enhanced}), leaving only a \PaperValue{v8a09eab4d99fd7df}~pp descriptive difference for BIOT and
no controlled foundation-model advantage. Second, REVE is numerically highest on TUAB, but its
\PaperValue{vcbbc4d939e299c96}-epoch sample differs from the \PaperValue{ve54ede0b638c1a02}-epoch sample used by the other rows, and pretraining overlap
is confirmed for every encoder; no TUAB model margin is therefore estimated. Third, on
the external CAUEEG cohort the ordering inverts: the handcrafted column is highest and every encoder
sits below it. Fourth, encoder coverage is not a full grid, so a column mean would be computed over
different task sets per model and is not reported.

\noindent\textbf{The study evaluates claim robustness, not an overall model ranking.} Because the
tasks differ in cohorts, outcomes, preprocessing, representation construction, and validation units,
the study does not seek an overall model ranking. Instead, it evaluates whether apparent advantages
within each task remain credible after targeted controls. Beyond the provenance warnings above, the
primary CAUEEG analysis uses recording-level folds while the others use subject-disjoint folds, and
CHB-MIT uses 512-dimensional mean pooling, the ds004504 REVE analyses use the historical \PaperValue{v188e6eefc04fa197}-dimensional flat-LayerNorm representation, and TUAB uses token flattening. We therefore restrict ourselves to three descriptive observations. \emph{Task pattern:} On CHB-MIT and AD versus FTD, pretrained REVE is above matched fully random initialization, whereas on CAUEEG it is below; these directions are pipeline-specific. The AD-versus-HC mean is matched by stronger classical comparators (Section~\ref{sec:enhanced}). On AD versus FTD REVE ($\PaperValue{v9c0f67aae64c8b39}\%$) and BIOT-bipolar16 ($\PaperValue{v936f50c3c6559f35}\%$) exceed the classical mean ($\PaperValue{v5bef026461f3d681}\%$), by $\PaperValue{v4068a6ccfcce2bf7}$ and $\PaperValue{v5a9d51e7b1ab59f4}$~pp respectively, while LaBraM, CBraMod and EEGMamba fall below it; paired control intervals for both positive margins cross zero (Section~\ref{sec:randominit}). \emph{Channel-count
hypothesis:} no evaluated encoder has a mean AUROC above the simple classical mean on \PaperValue{v37567c5e47c98348}-channel
Sleep-EDF, whereas CHB-MIT uses \PaperValue{v8c87773002afea40} channels, but task, dataset, montage, and label construction are
confounded, so a matched channel-ablation experiment would be required before attributing this to
spatial coverage. \emph{Model-size hypothesis:} REVE has the highest numerical mean on TUAB, but
uses a separate epoch sample and overlap is confirmed for all five encoders; moreover, the benchmark varies parameter count together with
architecture, objective, corpus, exposure, and preprocessing compatibility, so it cannot establish
that larger models or corpora cause better clinical discrimination. The CHB-MIT row of
Table~\ref{tab:benchmark} motivates the task-specific controls below.

\noindent\textbf{Task construction on CHB-MIT.}\label{sec:chbmit_ictal} Because CHB-MIT carries the
paper's clearest pretrained-versus-random separation, its task definition warrants statement in the main text.
Positive epochs overlap an annotated seizure by $\geq50\%$; negative epochs are interictal segments
drawn from the \emph{same seizure-containing recordings}, so the task is within-session ictal-versus-interictal discrimination generalized across held-out patients. An independent sensitivity implementation gives $\PaperValue{v811a2c36bad919b3}$ both without and with a $\pm30$\,s peri-ictal guard band, $\PaperValue{v010139c57367b67f}$ when negatives are drawn from all available files, and $\PaperValue{vebd7d8a600e609f8}$ after excluding chb12. The primary run is $\PaperValue{va964875514f7949e}$; the small offset reflects independently extracted epochs and different negative identities under the same fixed 1:3 ictal:interictal sampling ratio, not a change in sample count. These checks reduce concern about recording-session and near-onset confounding. REVE exceeds the simple classical baseline by $+\PaperValue{v96a08164ca73fe56}$~pp (Wilcoxon $p{=}\PaperValue{v6049160f3660ec19}$, $d{=}\PaperValue{vbcfa8f09aa392952}$, $n{=}\PaperValue{v089d14b5891f734c}$). Two caveats bound that figure: LOSO training sets overlap, so the fits are not independent, and the amplitude-normalized input omits an amplitude-aware handcrafted comparator. The result does not license sample-precise onset localization or treatment-response prediction. The probe configuration is likewise not driving the estimate: on the 512-dimensional mean-pooled representation the in-fold PCA sweep is flat, giving $\PaperValue{v546ff21179d07226}$ at 50 dimensions, $\PaperValue{v9f3e4a737fbd4f20}$ at 100, $\PaperValue{va5ff1f382f6b4e18}$ at 200, and $\PaperValue{v59ed225df1912249}$ at the full 512, so the selected PCA-200 reduction does not materially change the result relative to the full representation. Exact negative-pool results and the dimensionality and probe-regularisation sensitivities are reported in Supplementary Material, Section~S4 (``CHB-MIT task construction and robustness'').

\noindent\textbf{Subject-level aggregation on ds004504.}\label{sec:ad_band_ablation} Aggregating epoch
probabilities to one score per subject raises AUROC for Classical and REVE on both ds004504 contrasts.
It preserves the REVE-over-Classical ordering on AD versus FTD but reverses the epoch-level ordering
on AD versus HC, where Classical is numerically higher. These changes show that epoch- and
subject-level AUROC are distinct estimands; fold dispersion remains descriptive. Supplementary
Material, Section~S5 (``Subject-level aggregation on ds004504''; Table~S6) reports the exact AUROCs,
changes from the epoch-level estimates, and retained fold SDs.

\subsection{The Negative-Control Suite}
\label{sec:controls}

The estimates in Table~\ref{tab:benchmark} are the input to this section, not its conclusion. Four
controls are applied, each targeting a different alternative explanation for an apparent gain:
a weak comparator (Section~\ref{sec:enhanced}), the architecture rather than the pretrained weights
(Section~\ref{sec:randominit}), probe capacity rather than label structure
(Section~\ref{sec:labelperm}), and label-specific rather than label-independent adaptation
(Section~\ref{sec:lora}). The controls are targeted rather than exhaustive. Control~1 compares the encoders against the handcrafted families within its five completed tasks; comparator adequacy was not adjudicated on TUAB. Controls~2 and~3 target task-specific REVE and BIOT claims, whereas Control~4 is REVE-only. The task coverage and analysis unit are stated in each control subsection.

\subsubsection{Control 1: Stronger handcrafted comparators}
\label{sec:enhanced}

The first control asks whether an apparent encoder advantage survives a comparator stronger than
band power. The current enhanced handcrafted family is evaluated on four benchmark tasks and CAUEEG; TUAB is excluded because its retained caches cannot support a protocol-matched rerun. The auxiliary ds003490 dataset carries no task and is
excluded by construction. Each included task keeps its own specified probe, folds, and epoch definition, so
Table~\ref{tab:enhanced} is a within-task simple-versus-enhanced contrast rather than a cross-task
ranking. The encoder with the highest descriptive AUROC among those provenance-eligible is shown for each task. Following success
condition~(i) of Section~\ref{sec:attribution_rubric}, the final column subtracts the stronger of the
simple and enhanced handcrafted families under the matched logistic probe; the simple family is stronger on AD versus FTD,
Sleep-EDF, and CAUEEG. Using the enhanced family unconditionally would create apparent encoder
gains on rows where it underperforms the simple family. Both delta columns use unrounded values and
therefore need not equal differences calculated from the rounded cells shown.

\begin{table}[H]
\centering
\scriptsize
\setlength{\tabcolsep}{1.3pt}
\caption{Control~1 matched-logistic AUROCs on four benchmark tasks and CAUEEG; TUAB was not evaluated. Differences are in percentage points.}
\label{tab:enhanced}
\begin{tabular}{@{}lccccc@{}}
\toprule
\textbf{Task} & \shortstack{\textbf{Simple}\\\textbf{AUROC (\%)}} & \shortstack{\textbf{Enhanced}\\\textbf{AUROC (\%)}} & \shortstack{\textbf{Enh. $-$ simple}\\\textbf{(pp)}} & \shortstack{\textbf{Eligible encoder}\\\textbf{AUROC (\%)}} & \shortstack{\textbf{Encoder $-$ best}\\\textbf{handcrafted (pp)}} \\
\midrule
Ictal detect (CHB-MIT)$^{\dagger}$ & \PaperValue{v8f3e52e58a29c618}\% & \PaperValue{v610bf82ff5236c83}\% & $+\PaperValue{vf905612104b49878}$ & REVE \PaperValue{v1983b53706abb3bb} & $+\PaperValue{v1ee1de8a99473f93}$ \\
AD vs Control$^\|$ & \PaperValue{ve8907089e5fdcc61}\% & \PaperValue{v45a4ef01b74d8208}\% & $+\PaperValue{vff2b6b82a8c3e683}$ & BIOT-bipolar16 \PaperValue{vb19d0558443f5600} & $+\PaperValue{v4857c3a67042bd6b}$ \\
AD vs FTD$^{\ast\ast}$ & \PaperValue{v5bef026461f3d681}\% & \PaperValue{vcc64374fac1bbeb9}\% & $-\PaperValue{v47ad2d6406ec64ce}$ & REVE \PaperValue{v9c0f67aae64c8b39} & $+\PaperValue{vda8a8a64d03d9a6b}$ \\
Wake/Sleep (Sleep-EDF)$^\S$ & \PaperValue{v780df097635c7608}\% & \PaperValue{v26cd9b1c2602fefe}\% & $-\PaperValue{v5c83af940374dbbf}$ & CBraMod \PaperValue{v94496782c36c58cc} & $-\PaperValue{ve7beb0d75f9145b2}$ \\
N/MCI/dementia (CAUEEG)$^\P$ & \PaperValue{v6811171afff96c88}\% & \PaperValue{v8c9ea0ac636080fb}\% & $-\PaperValue{v88a8596624cae846}$ & BIOT-bipolar16 \PaperValue{ve7e6246d845f1eb4} & $-\PaperValue{vb2f9e30dc909bef7}$ \\
\bottomrule
\multicolumn{6}{@{}p{\textwidth}@{}}{\footnotesize \emph{Provenance eligibility.} An encoder is eligible only if published pretraining inventories show no confirmed evaluation-cohort or close distributional exposure. $^{\dagger}$BIOT-bipolar16 was pretrained on CHB-MIT; REVE is the highest provenance-clean encoder. On Sleep-EDF, sleep-domain exposure excludes LaBraM and EEGMamba; BIOT is unavailable, and CBraMod has the highest estimate among the remaining eligible encoders. $^\|$AD versus Control values are notch-filtered. $^\S$Sleep-EDF uses matched subjects, epochs, folds, and $C=1$; a separate feature-count sensitivity is descriptive. $^{\ast\ast}$AD versus FTD uses \PaperValue{va341d39cd19b0dd6} epochs from \PaperValue{v8b57fcf7d74c901f} subjects. $^\P$CAUEEG uses the prespecified LogReg comparison on identical folds; Supplementary Material, Section~S3.1 reports the complete three-probe matrix. The eligible encoder is selected descriptively without model-selection-adjusted uncertainty. The matched-logistic CHB-MIT margin in this table is distinct from the post-hoc enhanced-GBM sensitivity reported below.}
\end{tabular}
\end{table}

Table~\ref{tab:enhanced} shows an increase from simple to enhanced classical features on two tasks:
CHB-MIT (\PaperValue{v8f3e52e58a29c618}\% to \PaperValue{v610bf82ff5236c83}\%) and notch-filtered AD versus Control
(\PaperValue{ve8907089e5fdcc61}\% to \PaperValue{v45a4ef01b74d8208}\%). On AD versus Control, enhanced classical exceeds REVE
(\PaperValue{v45a4ef01b74d8208}\% vs.\ \PaperValue{v6c4d910ddadcf793}\%), while the descriptively selected eligible BIOT-bipolar16 estimate is
\PaperValue{v8a09eab4d99fd7df}~pp higher still. On the other three tasks the enhanced mean is lower than the simple mean.
Using the stronger matched-logistic handcrafted family on each row, REVE remains \PaperValue{v1ee1de8a99473f93}~pp higher on CHB-MIT and \PaperValue{vda8a8a64d03d9a6b}~pp higher on AD versus FTD,
whereas CBraMod is \PaperValue{ve7beb0d75f9145b2}~pp lower on Sleep-EDF and BIOT-bipolar16 is \PaperValue{vb2f9e30dc909bef7}~pp lower on CAUEEG. The CHB-MIT, Sleep-EDF, and CAUEEG matched-logistic margins are descriptive point estimates without model-difference intervals. For AD versus FTD, the separate primary-pipeline REVE--simple comparison has a paired CI of $[-\PaperValue{vf405fe8ac2eac9d6},\PaperValue{v9c8f20dc41e5355f}]$~pp, which crosses zero (Section~\ref{sec:randominit}).

\noindent\textbf{Comparator choice can change a model-specific conclusion.} On Sleep-EDF,
provenance-eligible CBraMod (\PaperValue{v94496782c36c58cc}\%) is above the \PaperValue{v26cd9b1c2602fefe}\% enhanced mean while remaining below the
\PaperValue{ve2ca19956fedce4f}\% simple mean. On AD versus HC, the same reversal holds for the prespecified REVE comparison:
REVE (\PaperValue{v6c4d910ddadcf793}\%) is above the \PaperValue{ve8907089e5fdcc61}\% simple mean but below the \PaperValue{v45a4ef01b74d8208}\% enhanced mean. It does not hold
for the descriptively selected best eligible encoder, BIOT-bipolar16 (\PaperValue{vb19d0558443f5600}\%), which differs from
the enhanced mean by only \PaperValue{v8a09eab4d99fd7df}~pp. The Sleep-EDF and REVE AD-versus-HC comparisons remain descriptive means without model-difference intervals. For BIOT versus enhanced handcrafted features, the conditional paired pooled-OOF interval is $[-\PaperValue{v5c915601e2925637},\PaperValue{vc9df4ac5d2795b7a}]$~pp and crosses zero. Every point margin is smaller than the corresponding fold SD. The result is therefore sensitivity of
the comparison to the chosen model and handcrafted family, not evidence of an advantage in either
direction.

To further test whether nonlinear classifiers close the gap, we evaluated task-specific, post-hoc
GradientBoosting and MLP configurations. AD versus HC used balanced training-sample weights for GradientBoosting and a two-hidden-layer MLP with early stopping on an epoch-random subset of each outer training fold; CHB-MIT used an unweighted GradientBoosting fit and a one-hidden-layer MLP. These single-seed configurations are exploratory rather than a common tuned estimator. On Alzheimer's detection the observed mean gap closes, and not only through classifier capacity: the
linear enhanced family already reaches \PaperValue{v45a4ef01b74d8208}\%, \PaperValue{v650ae0ed5bcd6294}\,pp above notch-filtered REVE (\PaperValue{v6c4d910ddadcf793}\%), and an
MLP on the full \PaperValue{v2c6439cdc1da1ac5}-feature simple handcrafted family reaches $\PaperValue{vbbe1e3b71628f266} \pm \PaperValue{v5bbcbf2399f05dcc}$\% AUROC across folds, \PaperValue{v881cf29a70ff4195}\,pp above it. Its epoch-random inner validation is not participant-grouped, and seed stability is untested. Subject-level AUROC
is likewise not higher for REVE (\PaperValue{v8d68854096981149}\% vs.\ \PaperValue{v2c7ee9ace3685a7e}\% Classical; Section~\ref{sec:ad_band_ablation}).
An exploratory within-REVE frequency-band sensitivity is reported in Supplementary Material,
Section~S6 (``Exploratory REVE frequency-band ablation''); it uses $C{=}1$ and does not alter
the matched encoder-versus-classical comparison above.
On CHB-MIT, enhanced-GBM reaches $\PaperValue{v03ad6fa66a566705}$ versus REVE $\PaperValue{vb370fc4c5a7faa29}$. The paired equal-subject difference is $+\PaperValue{vfc5dd21abc2243b1}$~pp with a 95\% held-out-subject bootstrap CI of $[-\PaperValue{v95de63f94c69e163},\PaperValue{ve3df06a4e857ec2a}]$~pp and conditional exact sign-flip $p{=}\PaperValue{v19374cb3dd93f0e0}$. The interval crosses zero; LOSO training sets overlap, enhanced GBM was selected post hoc, and the normalized-input pipeline cannot reconstruct an amplitude-aware comparator because raw EDF waveforms and pre-normalization scales are unavailable in the retained materials.

\subsubsection{Control 2: Matched fully random initialization}
\label{sec:randominit}

The second control asks whether pretrained weights, as opposed to the architecture and probe, carry the signal. REVE was fully instantiated from its pinned architecture for each random seed; every audited encoder tensor differed from the checkpoint. BIOT was likewise instantiated from its pinned architecture, with zero ordinary trainable tensors identical to the checkpoint; its deterministic non-trainable channel-index tensor was excluded from that count. Table~\ref{tab:randominit} reports the completed model-specific comparisons.

\begin{table}[H]
\centering
\caption{Control 2: matched fully-random-initialization results. Values are AUROC; $\Delta$ is pretrained minus fully random. The raw-signal random-feature reference was run on CHB-MIT only.}
\label{tab:randominit}
\scriptsize
\setlength{\tabcolsep}{3pt}
\begin{tabular*}{\textwidth}{@{\extracolsep{\fill}}>{\raggedright\arraybackslash}p{2.8cm}cccc>{\raggedright\arraybackslash}p{4.0cm}@{}}
\toprule
\textbf{Task} & \textbf{Pretrained} & \shortstack{\textbf{Fully}\\\textbf{random}} & \shortstack{\textbf{Raw random}\\\textbf{features}} & \shortstack{\textbf{$\Delta$}\\\textbf{(pp)}} & \textbf{Retained comparison unit} \\
\midrule
CHB-MIT & $\PaperValue{v6dd95784b595c020}$ & $\PaperValue{vea7ad4b657e4f2d6}$ & $\PaperValue{v0a71f280022bee49}$ & $+\PaperValue{ve88e1fe14712ab15}$ & \PaperValue{v8e128931181e4de9} fold scores per seed \\
CAUEEG & $\PaperValue{vf8e66e8a6ceffaac}$ & $\PaperValue{v420e51509322903f}$ & --- & $-\PaperValue{v6b8f67b2ee24abd4}$ & \PaperValue{v4bead4fac078cdf3} paired run means \\
AD vs. FTD (REVE) & $\PaperValue{vbdd01c1fb5051cbe}$ & $\PaperValue{vfe75b9de5087c30e}$ & --- & $+\PaperValue{v181e916396b84973}$ & five fixed folds, three random seeds \\
AD vs. HC (BIOT) & $\PaperValue{v6a6b32e1f7d0f2b5}$ & $\PaperValue{v7b629439ea07f010}$ & --- & $+\PaperValue{vcf5c85c16b40ab3d}$ & five fixed folds, one random seed \\
AD vs. FTD (BIOT) & $\PaperValue{v5743c89a1114e349}$ & $\PaperValue{v76bbf783959c8c12}$ & --- & $+\PaperValue{v2a566e725669ebee}$ & five fixed folds, one random seed \\
\bottomrule
\end{tabular*}
\par\smallskip{\scriptsize\raggedright CHB-MIT and CAUEEG REVE rows use three random seeds; AD-versus-FTD REVE uses three random seeds on the primary fixed folds; each BIOT row uses one fully random seed. Paired uncertainty for ds004504 is reported in the text because its bootstrap estimand differs from the canonical mean-fold delta. All point contrasts remain conditional on the stated architecture, representation, probe, and folds.\par}
\end{table}

On CHB-MIT, matched fully randomly initialized REVE encoders with seeds \PaperValue{vd6052d609c05a63e}, \PaperValue{vf4de53a42934ff68}, and \PaperValue{v92029fbf6cc8451b} reach mean-fold
AUROCs of $\PaperValue{v7e372fd83049d9de}$, $\PaperValue{veea8976c8b90d9bd}$, and $\PaperValue{vd72c6bcd4096b637}$, respectively (grand mean $\PaperValue{vea7ad4b657e4f2d6}$), compared with
$\PaperValue{v6dd95784b595c020}$ for pretrained REVE. A Gaussian random projection of the normalized raw epoch signal
reaches $\PaperValue{v0a71f280022bee49}$ across three constructions. Thus the descriptive pretrained-minus-fully-random
difference is $+\PaperValue{ve88e1fe14712ab15}$~pp, while the raw-signal reference is near chance. Seed-specific held-out-subject
fold scores are retained, but the LOSO training sets overlap and the same held-out subjects recur across initialization seeds;
we therefore report no inferential interval or test. Three Gaussian constructions also do not bound what other raw-signal features could extract.
Within those limits, the result supports a pretrained-weight contribution under the evaluated
CHB-MIT pipeline rather than an architecture-only or probe-only reading.

On CAUEEG, every pretrained/fully-random pair uses the same clean 19-channel inputs, split, and
StandardScaler--PCA-50--SVM-RBF probe on the primary $\PaperValue{va0d1ec09c13cec9d}$-dimensional flattened-token,
whole-vector-LayerNorm recording representation. A six-record raw-EDF check spanning all three
classes reproduced the retained primary vectors at \texttt{rtol}/\texttt{atol} $10^{-4}$.
Across paired initialization/split states \PaperValue{vf6bf55f887680628}/\PaperValue{v9c86eb233cbc6ecb}, \PaperValue{vfcb74078420c5ad6}/\PaperValue{vcb0e73537a69568e}, and \PaperValue{v0c7055b3ffd56f85}/\PaperValue{vd13157aa2eaff822}, pretrained
run means are $\PaperValue{vc0a8a954f825e359}$, $\PaperValue{veca973b0ef491ef6}$, and $\PaperValue{v60d797a1ce397fa6}$, whereas the matched fully-random means are $\PaperValue{vecc2f05045986fc1}$,
$\PaperValue{vb923d3d63229a821}$, and $\PaperValue{v20a828c3dacaa344}$. Fully random initialization is therefore higher by $\PaperValue{v061e2cacf78bd76c}$, $\PaperValue{v3c8b7e4173713a09}$, and
$\PaperValue{vd2f673b26fed6ffb}$~pp, yielding grand means of $\PaperValue{vf8e66e8a6ceffaac}$ and $\PaperValue{v420e51509322903f}$. The direction is consistent across the
three runs, but the same \PaperValue{vc3f761bbb7600b12} recordings recur and test sets overlap across split states, so the
run means are not independent replications and no inferential $p$-value is claimed. The opposing
CHB-MIT and CAUEEG directions show that the pretrained-weight contribution is task-specific; they
do not support a general encoder effect in either direction.

On AD versus FTD, pretrained REVE reaches $\PaperValue{vbdd01c1fb5051cbe}$ and simple classical reaches $\PaperValue{v4d41ce14edd9afcf}$. Three fully random REVE encoders yield a grand mean of $\PaperValue{vfe75b9de5087c30e}$, a $+\PaperValue{v181e916396b84973}$~pp pretrained-minus-random separation. Every epoch representation uses the \PaperValue{v188e6eefc04fa197}-dimensional flat-LayerNorm contract and the same fixed five folds, fold-local scaling, and balanced logistic probe. The paired within-fold subject-cluster bootstrap for the REVE-minus-simple mean-fold difference is $+\PaperValue{v122f51ecfee0022f}$~pp with 95\% CI $[-\PaperValue{vf405fe8ac2eac9d6},\PaperValue{v9c8f20dc41e5355f}]$~pp. The interval crosses zero, so the positive point margin is not resolved despite the descriptive pretrained-versus-random separation.

For BIOT, the AD-versus-HC pretrained and single-seed fully-random means are $\PaperValue{v6a6b32e1f7d0f2b5}$ and $\PaperValue{v7b629439ea07f010}$; the paired participant-cluster interval for pretrained minus enhanced handcrafted is $[-\PaperValue{v5c915601e2925637},\PaperValue{vc9df4ac5d2795b7a}]$~pp and for pretrained minus random is $[-\PaperValue{vcf3c5516f2bc4883},\PaperValue{v87e7788a9c1fb692}]$~pp. On AD versus FTD, the corresponding means are $\PaperValue{v5743c89a1114e349}$ and $\PaperValue{v76bbf783959c8c12}$, with intervals $[-\PaperValue{v32cdb6a650eac840},\PaperValue{veb5cb2315c385ddf}]$~pp against simple handcrafted and $[-\PaperValue{vf369e6b562afdcd0},\PaperValue{vc669d37f6b83ba1f}]$~pp against random. These diagnosis-stratified participant-cluster intervals target pooled out-of-fold epoch-AUROC differences conditional on fixed predictions; they do not refit models or include comparator-selection or random-seed variance. All four cross zero, so neither the tiny AD-versus-HC margin nor the AD-versus-FTD margin supports resolved BIOT attribution.

\subsubsection{Control 3: Label permutation}
\label{sec:labelperm}

The third control asks whether a probe estimate reflects label structure or merely probe capacity on
high-dimensional embeddings. Labels are permuted and the identical probe is refit.
Table~\ref{tab:labelperm} reports every completed diagnostic label-permutation analysis and its
task-specific permutation scheme.

\begin{table}[H]
\centering
\caption{Control 3: model-specific label-permutation analyses. Observed and null AUROCs belong to the stated analysis unit and probe.}
\label{tab:labelperm}
\scriptsize
\setlength{\tabcolsep}{3.5pt}
\begin{tabular}{@{}lcccc>{\raggedright\arraybackslash}p{5.3cm}@{}}
\toprule
\textbf{Task} & \textbf{Observed} & \shortstack{\textbf{Null mean}\\$\boldsymbol{\pm}$\textbf{ SD}} & \textbf{Shuffles} & \textbf{Nominal $p$} & \textbf{Permutation and probe scope} \\
\midrule
CHB-MIT (REVE) & $\PaperValue{v6dd95784b595c020}$ & $\PaperValue{v9521ad1795eb7dc7} \pm \PaperValue{v5bba4711cb34a295}$ & \PaperValue{v815a3ee09aaf11c5} & $\PaperValue{v2d9f829bf5835570}$ & Within-subject shuffling; exact headline in-fold pipeline \\
AD/HC (BIOT) & $\PaperValue{v6a6b32e1f7d0f2b5}$ & $\PaperValue{va82a050cedcdcc24} \pm \PaperValue{va5132419b63691a3}$ & \PaperValue{v815a3ee09aaf11c5} & $\PaperValue{vda05124b7f49ccbf}$ & Participant labels; primary epoch pipeline, fixed folds \\
AD/HC (REVE) & $\PaperValue{v9c28514f5e8d4c99}$ & $\PaperValue{v02ded317a0df5421} \pm \PaperValue{v929e0bb0af4b5d3c}$ & \PaperValue{v68706bc58b0f31da} & $\PaperValue{vda95aa60647b6107}$ & Participant labels; direct subject-mean probe \\
AD/FTD (BIOT) & $\PaperValue{v5743c89a1114e349}$ & $\PaperValue{v6ef6c05958f1383d} \pm \PaperValue{v303bc19f9d64eedb}$ & \PaperValue{v815a3ee09aaf11c5} & $\PaperValue{vac8d68626b06048b}$ & Participant labels; primary epoch pipeline, fixed folds \\
AD/FTD (REVE) & $\PaperValue{v23d9be02bce3f7b1}$ & $\PaperValue{vc856505b802e9cda} \pm \PaperValue{vcd37033d8aead606}$ & \PaperValue{v815a3ee09aaf11c5} & $\PaperValue{vaccfdb0156dd60be}$ & Participant labels; direct subject-mean probe \\
AD/FTD/HC (REVE) & $\PaperValue{v7657968fc3d4d561}$ & $\PaperValue{v1879a2f5847674db} \pm \PaperValue{v85df81828ded8335}$ & \PaperValue{v815a3ee09aaf11c5} & $\PaperValue{v3d569d488d4c156b}$ & Participant labels; direct three-way subject-mean probe \\
\bottomrule
\multicolumn{6}{@{}p{0.95\textwidth}@{}}{\vspace{2pt}\scriptsize Every row reuses precomputed folds and applies the finite-randomization correction. CHB-MIT refits PCA-200, scaling, and logistic regression inside every LOSO training fold. BIOT rows retain the primary epoch-level pipelines; REVE ds004504 rows use one direct \PaperValue{v188e6eefc04fa197}-dimensional subject-mean vector per participant. The analyses test label dependence for their stated probes, not between-model effects or freedom from other confounding.}
\end{tabular}
\end{table}

On CHB-MIT, the label-permutation analysis now matches the headline pipeline exactly: within every fixed held-out-subject fold it fits PCA-200 on training embeddings, then fits the scaler and balanced $C{=}1$ logistic probe on the training coordinates. Across \PaperValue{v815a3ee09aaf11c5} within-subject permutations, the null mean is $\PaperValue{v9521ad1795eb7dc7}\pm\PaperValue{v5bba4711cb34a295}$ versus the reproduced observed mean of $\PaperValue{v6dd95784b595c020}$; zero permutations meet or exceed the observed value, giving $p{=}\PaperValue{v2d9f829bf5835570}$. This establishes label dependence for the exact headline representation and probe, not a model-comparison or population-level causal result.

The BIOT ds004504 permutations follow each primary epoch-level pipeline: one diagnosis is permuted per participant, broadcast to all of that participant's epochs, and evaluated on the fixed folds. AD/HC yields $\PaperValue{v6a6b32e1f7d0f2b5}$ versus $\PaperValue{va82a050cedcdcc24}\pm\PaperValue{va5132419b63691a3}$ ($p{=}\PaperValue{vda05124b7f49ccbf}$), and AD/FTD yields $\PaperValue{v5743c89a1114e349}$ versus $\PaperValue{v6ef6c05958f1383d}\pm\PaperValue{v303bc19f9d64eedb}$ ($p{=}\PaperValue{vac8d68626b06048b}$). Thus both BIOT probes depend on the observed participant diagnoses, although that does not resolve their pretrained-versus-comparator or pretrained-versus-random margins.

The REVE ds004504 sensitivities instead average the same historical flat-LayerNorm epoch vectors within participant and fit a direct fold-local scaled probe. AD/HC and three-way AD/FTD/HC exceed their permutation nulls ($p{=}\PaperValue{vda95aa60647b6107}$ each). Binary AD/FTD yields $\PaperValue{v23d9be02bce3f7b1}$ versus $\PaperValue{vc856505b802e9cda}\pm\PaperValue{vcd37033d8aead606}$, with $p{=}\PaperValue{vaccfdb0156dd60be}$; this subject-mean control does not establish label dependence at the conventional 0.05 level and is not the primary epoch-level estimand. No Control~3 result is reported for TUAB, Sleep-EDF, or CAUEEG. None of these randomization tests is a between-model comparison or a test against every leakage or confounding mechanism.

\subsubsection{Control 4: Scrambled-label adaptation}
\label{sec:lora}

The fourth control asks whether a gain from task-specific adaptation depends on the labels being
correct. We treat adaptation as a secondary sensitivity analysis because model, task, split, and
optimization recipe vary across configurations. LoRA~\cite{hu2022lora} inserts trainable low-rank
matrices while leaving the pretrained backbone frozen; these experiments do not support a general
sample-size, model-size, or deployment rule.

\noindent\textbf{CAUEEG adaptation stability comparisons.} On the binary task, the three-run correct-label mean (\PaperValue{v9a056397a7d45572}) is within $\PaperValue{v368139a4bff45772}$~pp of the five-run scrambled-label mean (\PaperValue{vf55a08d33c89b541}). The correct-label runs used up to 20 training epochs with early-stopping patience 5, whereas the scrambled-label runs used up to 10 epochs with patience 3; their seed sets are also not paired. This is therefore an unmatched descriptive sensitivity, not an equivalence or label-effect test. On the three-way task, the source-cap-matched correct-label run (\PaperValue{v8db5c489f89aa72c}) is within $\PaperValue{v8d266bb25d2ef367}$~pp of the five-run scrambled-label mean (\PaperValue{vc9e75c1e5a8373f1}), but the downstream SVM settings differ and the correct-label run did not retain an explicit training seed. That comparison likewise does not isolate label correctness.

Adapted arms read CAUEEG amplitudes in microvolts, whereas the clean frozen reference receives volt-scale values, and their probes differ. Adapted-versus-frozen contrasts therefore cannot be interpreted as adaptation effects. Supplementary Material, Section~S7 (``Adaptation protocol provenance and seed structure'') reports the complete results, source-cap and full-source distinction, seed structure, and protocol differences. Within those limits, the numerical similarity does not provide evidence that correct Western source labels improved CAUEEG performance, but no mechanism is identified.

\subsection{External-Cohort Stress Test (CAUEEG, Normal/MCI/Dementia)}
\label{sec:domain_generalization}

\begin{figure}[tbp]
\centering
\includegraphics[width=0.94\textwidth]{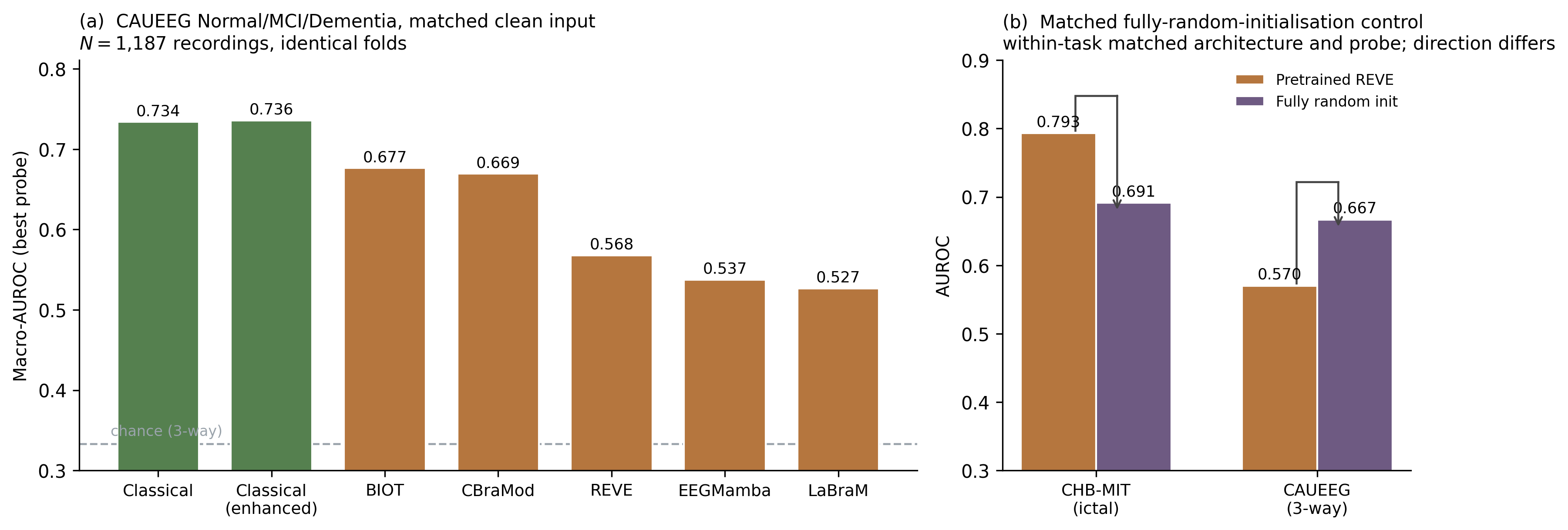}
\caption{The external-cohort stress test and its fully-random-initialisation control.
\textbf{(a)}~Best-probe macro-AUROC for every representation from the clean \PaperValue{v9ad2a8acbb26ddba}-channel CAUEEG
source montage under identical folds; BIOT uses the 16 upstream bipolar derivations constructible from it. The simple classical family is the headline comparator; the
enhanced family is shown as the comparator sensitivity and is $\PaperValue{v6f0f61a377b647ba}$~pp above it, which we do
not promote (Section~\ref{sec:enhanced}). All five encoders fall below both; relative to the simple
headline comparator, the deficits are $\PaperValue{vd48be45818bd4cba}$ to $\PaperValue{v07acc710b474d17e}$~pp.
\textbf{(b)}~The matched fully-random-initialisation control on the two tasks shown. Pretrained
REVE is above fully random initialization on CHB-MIT and below it on CAUEEG, so the observed contribution
of pretrained weights is task-specific rather than uniformly positive. Panel~(b) is descriptive:
fold positions are not paired across the repeated CAUEEG splits and no inferential $p$-value is claimed.}
\label{fig:caueeg}
\end{figure}

Figure~\ref{fig:caueeg} summarizes the matched clean-input comparison and the two-task
fully-random-initialization contrast.

\noindent\textbf{Matched clean-input comparison.} CAUEEG differs from the benchmark cohorts and exposes no patient identifier~\cite{kim2023caueeg}; the primary analysis is recording-level. Every representation starts from the same clean \PaperValue{v9ad2a8acbb26ddba}-channel montage, excluding EKG and Photic; BIOT receives the 16 upstream bipolar derivations constructible by named subtraction. All rows use identical five-fold partitions and LogReg, LDA, and SVM-RBF probes. The simple classical family reaches $\PaperValue{vb8c001cb226923a9}$ AUROC, versus BIOT $\PaperValue{v5d9aa6cbb1c6f7dc}$, CBraMod $\PaperValue{v9297353d3eab8509}$, REVE $\PaperValue{vf0606d3c224b784c}$, EEGMamba $\PaperValue{veda5a9d9896e7ed5}$, and LaBraM $\PaperValue{vd29f27e84a2ae739}$: a graded $\PaperValue{vd48be45818bd4cba}$--$\PaperValue{v07acc710b474d17e}$~pp deficit relative to the simple headline comparator. The enhanced SVM-RBF sensitivity is $\PaperValue{va4f8dff75a80ecd0}$, only $\PaperValue{v6f0f61a377b647ba}$~pp above simple SVM-RBF and without a difference interval; its matched LogReg estimate is lower ($\PaperValue{v053fbf33f66c2937}$ vs.\ $\PaperValue{v8789cb06f501dc6c}$). We therefore retain simple classical as the headline comparator and treat LDA/SVM-RBF and best-of-family summaries as descriptive sensitivities. Supplementary Material, Section~S3.1 (``Complete clean-input probe matrix'') is the authoritative full-precision matrix and includes fold SDs. The separate adaptation sensitivities and their protocol mismatches are reported in Section~\ref{sec:lora} and Supplementary Material, Section~S7.

\subsection{Linear Probes Perfectly Decode Dataset Membership}
\label{sec:identity}

\noindent\textbf{Dataset-identity probe.} All five encoders separate ds004504 from CAUEEG at AUROC $\PaperValue{vd0d193f424fcfe36}$ in full space and after in-fold PCA-50 (Table~\ref{tab:identity}). BIOT uses the 16 upstream bipolar derivations available in both cohorts; the other encoders use their specified \PaperValue{v9ad2a8acbb26ddba}-channel representations. Across \PaperValue{v6843f6d2515ba41c} label permutations the encoder means are $\PaperValue{v82c48b60d09b5c24}$--$\PaperValue{vce6a66b1022fbfde}$, showing that the probe does not produce ceiling values for arbitrary labels; ten balanced $\PaperValue{v1d845c6999c3e8eb}$-versus-$\PaperValue{v1d845c6999c3e8eb}$ draws remain at $\PaperValue{v8c7c2b99622623a6}$, ruling out class imbalance as the explanation. The REVE-only Western--Western ds004504--ds003490 sensitivity reaches $\PaperValue{v91c163ee262cb289}$ in full space and $\PaperValue{ve9e7ee15373fc2d2}$ after in-fold PCA-50. After restricting ds004504--CAUEEG inputs to \PaperValue{ve1fd0c6fac5ac184}--\PaperValue{vb9eed640f212fafe}\,Hz and z-scoring each epoch, REVE reaches $\PaperValue{vaca7da095d4a394c}$ in full space and $\PaperValue{v1bb724e9ce3cd967}$ after in-fold PCA-50. For REVE, near-ceiling decoding is therefore neither unique to the tested Western--Korean pair nor removed with line-frequency and absolute-scale cues. The result remains dataset membership: device, montage, preprocessing, population, and clinical composition vary together and cannot be separated causally. The two REVE-only extensions are reported in Supplementary Material, Section~S8 (``Supplemental dataset-identity controls'').

\begin{table}[H]
\centering
\caption{Dataset-identity decoding is a property of every evaluated encoder, not of REVE alone.
Values are mean five-fold AUROC for predicting which dataset produced a recording (ds004504
$n{=}\PaperValue{v8cc1fbf12d281798}$ versus CAUEEG $n{=}\PaperValue{v5071b3c9d6a5451f}$), under one probe applied to each encoder's frozen
specified representation (BIOT upstream bipolar16; \PaperValue{v9ad2a8acbb26ddba} channels for the other encoders). The two right-hand columns are controls on the ceiling itself: with
the dataset label permuted every encoder falls to chance, and on a balanced $\PaperValue{v1d845c6999c3e8eb}$-versus-$\PaperValue{v1d845c6999c3e8eb}$
subsample every encoder stays at ceiling. The label is dataset membership; site, device,
montage, preprocessing, population, and clinical composition are confounded within it.}
\label{tab:identity}
\footnotesize
\setlength{\tabcolsep}{6pt}
\begin{tabular}{@{}lccccc@{}}
\toprule
\textbf{Encoder} & \textbf{Dim.} & \textbf{Full space} & \textbf{PCA-50} & \textbf{Permuted label} & \textbf{Balanced \PaperValue{v1d845c6999c3e8eb} vs.\ \PaperValue{v1d845c6999c3e8eb}} \\
\midrule
REVE     & \PaperValue{v9a7b56a1817bc304} & $\PaperValue{v66d94c8443fb746d}$ & $\PaperValue{v277037d233fc1442}$ & $\PaperValue{vfeda25c57cf14cea} \pm \PaperValue{v0280292fe6c4e715}$ & $\PaperValue{v8c7c2b99622623a6}$ \\
BIOT     & \PaperValue{v46852ea18dccb863}      & $\PaperValue{vde9d729bb7152fbd}$ & $\PaperValue{vcaec29a078df7ebc}$ & $\PaperValue{vd04ca523a8326f16} \pm \PaperValue{vd6b8823cda21a851}$ & $\PaperValue{vcb589cdcf41ef0be}$ \\
CBraMod  & \PaperValue{vfab46c67ff68f0b5}      & $\PaperValue{v6c06cce3e0a50704}$ & $\PaperValue{v8cedc6c358f1628c}$ & $\PaperValue{vf6faed7a5fbbafe6} \pm \PaperValue{v55eedafc6a50cc33}$ & $\PaperValue{vd8a7795a4750560e}$ \\
EEGMamba & \PaperValue{v21a4d0b9b1ca4e3b}      & $\PaperValue{v4f37e47574d60b71}$ & $\PaperValue{va31d2d8d7fdc2ad8}$ & $\PaperValue{v98afc1e353c2581b} \pm \PaperValue{vb5ae71193d8cb527}$ & $\PaperValue{v4eac869a6372d098}$ \\
LaBraM   & \PaperValue{v68560cf0795c08cb}      & $\PaperValue{v646bd7f009f3f4fd}$ & $\PaperValue{v2f1ff596853dc118}$ & $\PaperValue{ve65202e34fc89947} \pm \PaperValue{v90b2db9ffaf5bc7a}$ & $\PaperValue{v57f1ab098fdca56f}$ \\
\bottomrule
\multicolumn{6}{@{}p{0.95\textwidth}@{}}{\footnotesize Probe: per-feature standardisation, then
optional in-fold PCA-50, then $L_2$-regularised logistic regression with balanced class weights,
five stratified folds, seed \PaperValue{v9354aab0e2577398}. Permutation uses \PaperValue{v6843f6d2515ba41c} shuffles and the balanced control \PaperValue{v0e5c4529624f7407} draws,
both at PCA-50. Because all five encoders sit at $\PaperValue{vd0d193f424fcfe36}$, the probe supports a shared property of
the panel and cannot rank encoders. The Western--Western ds003490 pair and the \PaperValue{ve1fd0c6fac5ac184}--\PaperValue{vb9eed640f212fafe}\,Hz
z-scored sensitivity were run for REVE only.}
\end{tabular}
\end{table}

\subsection{Patient-Overlap Sensitivity}
\label{sec:overlap}

\noindent\textbf{Patient-overlap sensitivity.} In the public CAUEEG repository (\url{https://github.com/ipis-mjkim/caueeg-dataset}, accessed 6 August 2026), the authors' \texttt{dementia-no-overlap.json} task file provides the available patient-overlap annotation~\cite{kim2023caueeg}. It removes \PaperValue{v5819c9a7086222c3} of \PaperValue{vdc0745a0fd565316} held-out recordings while leaving the training split fixed. On the resulting annotated no-overlap held-out subset ($n{=}\PaperValue{v4cd043e2f21b8269}$ recordings), classical LogReg reaches $\PaperValue{v91a0efa5161478aa}$ (recording-bootstrap 95\% CI $[\PaperValue{v8c344acbb3e6c42b},\PaperValue{v523603f3ff2e9f55}]$) versus REVE LogReg $\PaperValue{v822c2b5c74b87d0e}$ ($[\PaperValue{v34432728441f3902},\PaperValue{ve78b44b95d50becb}]$). The classical--REVE gap is about five percentage points larger in the overlap-permitting fixed-split sensitivity than on the annotated no-overlap subset, so overlap may inflate the observed gap. This is not a mathematical bound on the separate five-fold estimator: the bootstrap treats recordings as independent, general patient identifiers are unavailable, and repeat recordings wholly within training remain unquantified. The complete full-, annotated no-overlap, and overlap-only probe matrix is reported in Supplementary Material, Section~S3.2 (``Patient-overlap sensitivity'').

\subsection{Inference Status}
\label{sec:inference_status}

\noindent\textbf{Primary-task inference.} TUAB, Sleep-EDF, and the primary
CAUEEG five-fold matrix remain descriptive: they have point estimates and fold
variation, but no paired model-difference interval. The CAUEEG intervals in the
preceding subsection are marginal recording-bootstrap intervals from one fixed
annotated no-overlap split; they are not patient-level intervals or paired
between-model effects. The complete overlap matrix is reported in Supplementary
Material, Section~S3.2.

On CHB-MIT, the REVE-versus-simple comparison gives Wilcoxon
$W{=}\PaperValue{v82fcef5335dde81a}$, nominal
$p{=}\PaperValue{v6049160f3660ec19}$, and paired
$d{=}\PaperValue{vbcfa8f09aa392952}$ across
\PaperValue{v089d14b5891f734c} held-out-subject folds. Against the stronger
post-hoc enhanced GBM, the paired difference is
$+\PaperValue{vfc5dd21abc2243b1}$~pp with CI
$[-\PaperValue{v95de63f94c69e163},\PaperValue{ve3df06a4e857ec2a}]$~pp and
conditional sign-flip $p{=}\PaperValue{v19374cb3dd93f0e0}$. The interval crosses
zero; moreover, LOSO training sets overlap and the stronger comparator was
selected after inspection, so neither result is confirmatory.

For ds004504 AD versus HC, the marginal REVE and classical fold intervals do
not estimate a paired model difference. The conditional BIOT-minus-enhanced interval
$[-\PaperValue{v5c915601e2925637},\PaperValue{vc9df4ac5d2795b7a}]$~pp crosses
zero. For AD versus FTD, the REVE-minus-simple difference is
$+\PaperValue{v122f51ecfee0022f}$~pp with CI
$[-\PaperValue{vf405fe8ac2eac9d6},\PaperValue{v9c8f20dc41e5355f}]$~pp, and the
BIOT-minus-simple CI is
$[-\PaperValue{v32cdb6a650eac840},\PaperValue{veb5cb2315c385ddf}]$~pp; both
cross zero. The REVE interval resamples participants within fixed folds and
targets the mean-fold difference, whereas the BIOT intervals target pooled
out-of-fold epoch-AUROC differences conditional on fixed predictions. Neither
procedure refits models or incorporates comparator-selection or random-seed
uncertainty.

\noindent\textbf{Targeted controls.} The fully-random-initialization contrasts
in Table~\ref{tab:randominit} are model-specific; the ds004504 paired intervals
are conditional on fixed predictions and exclude random-seed variance. The
permutation analyses in Table~\ref{tab:labelperm} test label dependence within
their stated units and probes, not patient-level population effects or
between-model differences. CAUEEG random-initialization runs also reuse
recordings across split states. These controls therefore answer narrower
questions and do not confer confirmatory inference on the primary comparisons.
CHB-MIT robustness details are in Supplementary Material, Section~S4, and
adaptation-protocol provenance is in Section~S7.

\section{Discussion}

Across the six primary tasks, no encoder advantage satisfies all four attribution
conditions. Interpretation depends on pretraining provenance, comparator strength, probe
design, and analysis unit (Section~\ref{sec:inference_status}).

\subsection{Observed Advantages Depend on Task, Provenance, and Comparator}

On TUAB, REVE has the highest numerical mean, but it uses a separate \PaperValue{vcbbc4d939e299c96}-epoch sample while the other reported rows use \PaperValue{ve54ede0b638c1a02} epochs; Table~\ref{tab:benchmark} also documents confirmed TUH or adjacent-distribution exposure for every encoder. TUAB therefore supplies neither a matched model margin nor evidence that scale or transfer caused the numerical ordering, and comparator adequacy was not adjudicated with the current enhanced family. On AD versus HC, enhanced classical features and an exploratory MLP on the full simple handcrafted family have mean AUROCs that match or exceed REVE. BIOT is only $\PaperValue{v8a09eab4d99fd7df}$~pp above enhanced classical in the mean-fold summary, and its paired conditional interval crosses zero. On CHB-MIT, pretrained REVE is higher than fully random initialization and the tested normalized-input baselines, but the paired strongest-comparator interval crosses zero and an amplitude-aware comparator cannot be reconstructed from the retained normalized epochs.

The CAUEEG stress test sharpens that caution. Under the shared clean-input protocol, all five encoders range from $\PaperValue{vd29f27e84a2ae739}$ for LaBraM to $\PaperValue{v5d9aa6cbb1c6f7dc}$ for BIOT-bipolar16, below $\PaperValue{vb8c001cb226923a9}$ for simple classical features ($\PaperValue{va4f8dff75a80ecd0}$ for the enhanced sensitivity). In the repeated-split fully-random-initialization comparison, fully random initialization is descriptively higher in all three run means, which is directionally inconsistent with a benefit from REVE's pretrained weights under the evaluated probe; no inferential $p$-value is claimed because the runs reuse recordings. Linear probes decode dataset membership at AUROC $\PaperValue{vd0d193f424fcfe36}$ for all five encoders in both full space and after in-fold PCA-50, but this observation does not explain the clinical-task deficit causally: site, device, montage, preprocessing, population, and diagnosis composition are bundled. These results do not establish that greater pretraining scale or a favorable numerical position on another task yields better performance on this cohort.

\subsection{Channel Count as a Hypothesis}
\label{sec:channel_density}

None of the four encoders evaluated on Sleep-EDF has a mean AUROC above the classical mean on the \PaperValue{v37567c5e47c98348}-channel binary task; no paired model-comparison test was computed. Sparse spatial coverage and the binary coarsening of the sleep-stage labels are plausible contributors, but they are confounded in this single dataset and were not experimentally separated. The result therefore motivates, rather than establishes, a channel-density hypothesis. Testing that hypothesis requires a richer-montage sleep cohort with controlled within-recording channel ablation while holding the label construction fixed.

\subsection{Frozen Probing and Configuration-Specific Adaptation}

The divergence between cross-subject ictal detection and the other evaluated configurations suggests hypotheses about compatibility with frozen probing, but the benchmark does not isolate task type from dataset and representation:

\begin{itemize}[noitemsep]
    \item \textbf{Discrete events.} CHB-MIT ictal versus interictal discrimination is captured by frozen REVE and shows a descriptive pretrained-versus-fully-random-initialization separation; untested event types such as spike-wave discharges or sleep spindles are not implied.
    \item \textbf{Diffuse states.} Results are mixed: REVE leads descriptively on TUH normal/abnormal, but confirmed TUH distributional overlap prevents a transfer interpretation; stronger classical comparators close the AD-versus-HC mean gap; and every evaluated encoder is below simple classical on CAUEEG Normal/MCI/dementia. Medication effects were not evaluated.
    \item \textbf{Differential diagnosis.} On AD versus FTD, REVE ($\PaperValue{v9c0f67aae64c8b39}\%$) and BIOT-bipolar16 ($\PaperValue{v936f50c3c6559f35}\%$) are descriptively above simple Classical ($\PaperValue{v5bef026461f3d681}\%$), whereas LaBraM, CBraMod, and EEGMamba are below it. REVE is $\PaperValue{v181e916396b84973}$~pp above the three-seed fully-random grand mean, but its paired comparator CI $[-\PaperValue{vf405fe8ac2eac9d6},\PaperValue{v9c8f20dc41e5355f}]$~pp crosses zero and its subject-mean label permutation gives $p{=}\PaperValue{vaccfdb0156dd60be}$. BIOT is $\PaperValue{v2a566e725669ebee}$~pp above one fully-random seed and passes its participant-label permutation, but both paired conditional intervals cross zero. The controls therefore characterize the positive point estimates without establishing either encoder's advantage over handcrafted features.
\end{itemize}

LoRA sensitivities are reported in Section~\ref{sec:lora} and Supplementary Material, Section~S7. Both the binary and three-way correct-versus-scrambled comparisons are unmatched descriptive sensitivities because their training or downstream-probe protocols differ; neither is an equivalence test or a label-effect estimate.

\noindent\textbf{Adaptation remains configuration-specific.} The LoRA experiments vary model family, task, sample size, and training recipe simultaneously, so they do not establish universal sample-size or model-to-data thresholds. They show only that adaptation can be unstable and that, under these settings, a claim of label-specific transfer requires a scrambled-label control.

\subsection{Limitations}
\label{sec:limitations}

\begin{enumerate}[noitemsep]
    \item \textbf{Diagnostic tasks only, on public or registration-gated data.} Every task here is retrospective diagnostic or state classification on third-party public or registration-gated cohorts. The protocol is not validated for, and these results say nothing about, treatment-response or pharmacodynamic biomarkers, which would require evaluation of within-subject change, test--retest reliability, and a defined context of use. Extending the control suite to longitudinal, outcome-labelled data is left to future work.
    \item \textbf{Subject-level metrics.} Subject aggregation was retained only for the two ds004504 contrasts and cannot be reconstructed for TUAB from the retained data. Its task-specific ordering changes demonstrate that epoch- and subject-level estimates are not interchangeable; they do not establish clinical reliability or cross-cohort stability (Section~\ref{sec:ad_band_ablation}; Supplementary Material, Section~S5, Table~S6).
    \item \textbf{Limited multi-class scope.} Beyond binary tasks (seizure vs.\ interictal, normal/abnormal, wake/sleep), the primary evaluation includes one three-way task (CAUEEG Normal/MCI/Dementia). A separate REVE-only Control~3 sensitivity uses three-way ds004504 AD/FTD/HC. Regression and severity-scoring tasks are not assessed.
    \item \textbf{Model coverage and pretraining exposure.} BIOT was evaluated with its bipolar16 input contract on CHB-MIT, both ds004504 contrasts, and CAUEEG. Sleep-EDF cannot supply the contract, and TUAB is not reported because the required named-channel input is unavailable. BIOT reaches \PaperValue{v6a420d0cc62641f0} on CHB-MIT but was pretrained on that dataset~\cite{yang2023biot}, so its estimate cannot distinguish memorization from distribution familiarity. REVE and LaBraM are descriptively similar (\PaperValue{vb370fc4c5a7faa29} and \PaperValue{v6810426a7cb2ffa4}); additional controls cover REVE on CHB-MIT, AD versus FTD, and CAUEEG, and BIOT on both ds004504 contrasts, not the complete model--task grid.
    \item \textbf{REVE sleep evaluation note.} REVE was evaluated on Sleep-EDF using compound channel names (Fpz-Cz, Pz-Oz) resolved to single electrodes (Fpz, Pz) for the REVE position bank. This coordinate substitution is a practical approximation. Evaluation on a separate sleep cohort with a full 10--20 montage and controlled channel ablation would provide a cleaner spatial-signal test.
    \item \textbf{REVE's gated license.} REVE's model-card access requires registration and agreement to responsible-use terms, limiting reproducibility for groups without access (\url{https://huggingface.co/brain-bzh/reve-base}, accessed 6 August 2026).
    \item \textbf{Position embedding interpolation for LaBraM.} The executed loader linearly interpolates LaBraM's 129-entry spatial position tensor to $N_{\mathrm{ch}}+1$ entries and its 16-entry temporal tensor to five positions for the 4-s inputs. These study-specific transformations may alter its representations; published LaBraM benchmarks use the native 128-channel configuration~\cite{jiang2024labram}.
    \item \textbf{EEGMamba implementation fidelity.} EEGMamba's CUDA-only Mamba2 selective scan was reimplemented in pure PyTorch for CPU/MPS execution. All evaluated EEGMamba estimates are conditional on this local port, whose numerical parity with the upstream CUDA kernel was not independently tested.
    \item \textbf{Consumer hardware execution and TUAB subsampling.} Computation was performed on one Mac M3 (16~GB) using CPU or MPS, which does not establish computational portability or clinical deployability. From \PaperValue{vd01f79f27741aa43} available TUAB epochs, Classical, LaBraM, EEGMamba, and CBraMod use a \PaperValue{ve54ede0b638c1a02}-epoch subject-capped sample, whereas REVE uses a separate \PaperValue{vcbbc4d939e299c96}-epoch global sample. Full-set encoder estimates were not run, and the REVE and non-REVE cells do not estimate matched model differences. A current-family enhanced comparator could not be reconstructed because the retained TUAB caches do not include the required raw signals and participant-group vector.
    \item \textbf{Task-specific extraction.} CHB-MIT uses 512-dimensional mean-pooled REVE embeddings, ds004504 uses the historical \PaperValue{v188e6eefc04fa197}-dimensional flat-LayerNorm representation, and TUAB uses token flattening. The REVE ds004504 label-permutation sensitivities average the flat-LayerNorm epoch vectors within participant; the primary ds004504 estimates and random-initialization control remain epoch-level. Extraction choices were not compared systematically across every task.
    \item \textbf{Limited statistical inference.} The paired REVE-versus-simple-classical CHB-MIT comparison yields nominal $p=\PaperValue{v6049160f3660ec19}$ and paired $d=\PaperValue{vbcfa8f09aa392952}$, but LOSO training sets overlap. Against enhanced GBM, the paired CI $[-\PaperValue{v95de63f94c69e163},\PaperValue{ve3df06a4e857ec2a}]$~pp crosses zero; the comparator was selected post hoc and no amplitude-aware comparator can be reconstructed. The ds004504 paired intervals also cross zero and are conditional on fixed folds or predictions; the BIOT random comparison has one seed. CAUEEG repeated splits reuse recordings. These comparisons are bounded sensitivities rather than confirmatory population tests.
    \item \textbf{Label-dependence analyses.} Control~3 covers exact-headline CHB-MIT REVE; primary-pipeline BIOT on both ds004504 contrasts; and subject-mean REVE on AD/HC, binary AD/FTD, and three-way AD/FTD/HC. The CHB-MIT and BIOT observed-label probes exceed their permutation nulls; the binary subject-mean REVE AD/FTD control does not establish label dependence at the conventional 0.05 level. No label-permutation result is reported for TUAB, Sleep-EDF, or CAUEEG. Each result is restricted to its stated unit, probe, and permutation scheme and is not a population test or model comparison (Section~\ref{sec:labelperm}).
    \item \textbf{CAUEEG metadata do not support patient-grouped cross-validation.} The primary CAUEEG folds are recording-level. On the $n=\PaperValue{v4cd043e2f21b8269}$ annotated no-overlap held-out recordings, clean classical LogReg has a higher point estimate than clean REVE LogReg ($\PaperValue{v91a0efa5161478aa}$ vs.\ $\PaperValue{v822c2b5c74b87d0e}$), with wide recording-bootstrap intervals. The gap was about five points larger in one overlap-permitting fixed split, so overlap may inflate it; this does not bound the separate five-fold estimator, and repeat recordings wholly within training remain unannotated. Validation on additional external cohorts with complete recordings and subject-level outcome labels remains necessary before any population-level claim is warranted.
\end{enumerate}

\subsection{Attribution Status Under the Interpretive Framework}
\label{sec:success_criterion}

No task supports all four conditions in Section~\ref{sec:attribution_rubric}. The sequential rule makes the conclusion direct: once a positive model-specific margin fails comparator uncertainty or provenance, later controls may explain the pipeline but cannot restore an attribution claim.

TUAB stops before the first complete gate because REVE and the non-REVE rows use unmatched samples; all encoders also have confirmed TUH or adjacent-distribution exposure. Sleep-EDF and CAUEEG stop at the comparator gate because every tested encoder mean is below the handcrafted comparator. On AD versus HC, REVE is matched or exceeded by stronger handcrafted configurations. BIOT has a tiny positive mean-fold margin, but its conditional paired comparator CI $[-\PaperValue{v5c915601e2925637},\PaperValue{vc9df4ac5d2795b7a}]$~pp crosses zero; its pretrained-versus-random interval also crosses zero even though its label-permutation test is positive.

On AD versus FTD, REVE and BIOT are above simple classical in the point estimates, but both stop at paired uncertainty. REVE's $+\PaperValue{v122f51ecfee0022f}$~pp difference has CI $[-\PaperValue{vf405fe8ac2eac9d6},\PaperValue{v9c8f20dc41e5355f}]$~pp; pretrained REVE is well above three fully-random seeds, whereas the subject-mean label audit gives $p{=}\PaperValue{vaccfdb0156dd60be}$. BIOT's paired comparator CI is $[-\PaperValue{v32cdb6a650eac840},\PaperValue{veb5cb2315c385ddf}]$~pp; it is descriptively above one random seed and its primary-pipeline label audit gives $p{=}\PaperValue{vac8d68626b06048b}$. Neither pattern establishes superiority over handcrafted features. The other three encoders fall below the simple classical mean.

CHB-MIT also stops at paired comparator uncertainty. REVE exceeds enhanced GBM by $\PaperValue{vfc5dd21abc2243b1}$~pp, but the paired held-out-subject CI $[-\PaperValue{v95de63f94c69e163},\PaperValue{ve3df06a4e857ec2a}]$~pp crosses zero (conditional sign-flip $p{=}\PaperValue{v19374cb3dd93f0e0}$). CHB-MIT is absent from REVE's published pretraining inventory, and both the matched random-initialization and exact headline label-permutation controls are positive, but those later gates cannot repair the unresolved strongest-comparator margin. The paired sensitivity is conditional on overlapping LOSO training sets and post-hoc comparator selection; an amplitude-aware comparator is blocked because the retained data contain only per-epoch normalized signals and no raw EDF waveforms or scale parameters.

\subsection{Implications for Clinical EEG AI}

REVE's CHB-MIT result is task-relevant separability, not onset localization or
external-population transfer. Paired uncertainty against the strongest tested comparator includes zero, and an amplitude-aware baseline cannot be reconstructed from the retained normalized data. More broadly, a clinical EEG foundation-model advantage
is a task-specific attribution claim. Evaluations should report pretraining provenance, use the
strongest task-appropriate handcrafted family, matched fully random initialization, label permutation,
and, for adaptation claims, a scrambled-label control. Model comparisons require paired
subject-level uncertainty when the analysis unit permits it. External-cohort evaluations should
use subject-grouped splits when identifiers are available, and dataset-membership decoding should
not be interpreted as diagnostic transfer. These are evaluation safeguards, not deployment rules.

\section{Conclusion}

\noindent Clinical EEG foundation-model claims should be evaluated as falsifiable
comparisons rather than a leaderboard. Under the controls applied here, three findings
remain. First, on the matched clean-input CAUEEG stress test every encoder falls below
the simple classical comparator by $\PaperValue{vd48be45818bd4cba}$--$\PaperValue{v07acc710b474d17e}$~pp; the enhanced-comparator sensitivity
preserves that all-encoder ordering, while the patient-overlap sensitivity separately
preserves only the classical-over-REVE ordering. This is one external
cohort with recording-level primary folds, not geographic replication. Second,
dataset identity is decodable at ceiling from all five encoders on the Western--Korean
pair and from REVE on the Western--Western pair. Because acquisition and cohort factors are bundled, this
warns about dataset structure but does not identify a causal site mechanism. Third,
CHB-MIT cross-subject ictal detection shows pretrained-weight and label dependence in the evaluated pipeline: pretrained REVE exceeds matched fully random initialization and raw-signal random features, and the exact headline pipeline exceeds its within-subject permutation null. Superiority over the strongest tested handcrafted comparator is unresolved because its $+\PaperValue{vfc5dd21abc2243b1}$~pp paired margin over enhanced GBM has CI $[-\PaperValue{v95de63f94c69e163},\PaperValue{ve3df06a4e857ec2a}]$~pp, and an amplitude-aware comparator cannot be recovered from the retained normalized data.

We therefore recommend model-compatible channel construction; subject-grouped
evaluation or an explicit leakage sensitivity; strong linear and nonlinear classical
comparators; matched fully random initialization; label permutation; a dataset-membership
probe; and scrambled-label adaptation controls when transfer learning is used. None
of the present tasks clears all four conditions in Section~\ref{sec:attribution_rubric},
so no general claim of disease-information transfer across clinical populations is
warranted.

\section*{Data and Code Availability}

CHB-MIT is available from PhysioNet~\cite{physionet2010chbmit}, and Sleep-EDF is available through PhysioNet~\cite{kemp2000sleep,goldberger2000physiobank}. TUH TUAB requires application to Temple University~\cite{obeid2016tuh,lopez2017tuab}. OpenNeuro ds004504 is publicly available from OpenNeuro~\cite{miltiadous2023alzheimer}. CAUEEG is distributed through the repository identified by its dataset publication~\cite{kim2023caueeg}. The auxiliary dataset used only for the dataset-identity control, ds003490~\cite{cavanagh2021ds003490}, is openly downloadable from OpenNeuro. The evaluated REVE checkpoint (\texttt{brain-bzh/reve-base}) requires gated Hugging Face access under its model-card terms (\url{https://huggingface.co/brain-bzh/reve-base}, accessed 6 August 2026). The evaluated LaBraM, EEGMamba (\texttt{weighting666/EEGMamba}), CBraMod, and BIOT (\texttt{braindecode/biot-pretrained-six-datasets-\PaperValue{vc9b151715e9af58f}chs}) checkpoints are identified in their respective model sources~\cite{jiang2024labram,wang2025eegmamba,wang2025cbramod,yang2023biot}; access terms are set by the hosting repositories. Earlier reproducibility materials are publicly available at \url{https://github.com/MarziehzZare/eeg-fm-negative-controls}. A manuscript-matched release of the analysis code, preprocessing configurations, aggregate evidence artifacts, canonical numerical registry and ledger, and evaluation protocols has been prepared and will be posted at the same URL before publication. Raw or processed EEG, participant-level embedding caches, and third-party model weights are not redistributed.

\section*{Acknowledgments}

We thank the authors of LaBraM, EEGMamba, CBraMod, REVE, and BIOT for releasing pretrained weights, and the Temple University Hospital EEG Corpus, PhysioNet, and OpenNeuro teams for maintaining open clinical EEG datasets. OpenAI Codex (GPT-5) and Anthropic Claude (Opus 5) were used under the author's direction for code development, manuscript drafting and restructuring, copy-editing, and adversarial consistency review. The author critically reviewed all resulting analyses and text and takes full responsibility for their accuracy, integrity, and originality. No AI-generated data were used, and no AI tool is listed as an author.

\section*{Funding statement}
This research received no external funding, and the work was self-funded by the corresponding author.

\section*{Conflict of interest statement}
The author has a potential commercial interest in EEG-based biomarker technology related to this work. The study used only third-party, de-identified, publicly available or registration-gated datasets; no proprietary data or commercial products were used or evaluated. The author declares no other competing interests.

\section*{Ethical statement}
This study is a secondary analysis of previously collected and de-identified EEG datasets. No new data were collected from human participants. Ethical approval for the original data collection was obtained by the respective data providers and is described in the original publications cited in Table~\ref{tab:datasets} and Section~\ref{sec:domain_generalization}. No additional ethical approval was required for this secondary analysis.

\section*{Author Contributions}
M.Z. is the sole author and conceived the study, designed the evaluation protocol and negative-control suite, implemented the analysis pipeline, performed all experiments and statistical analyses, and wrote and revised the manuscript.

\sloppy

\end{document}